\documentclass[preprint,12pt]{elsarticle}

\usepackage{amssymb}
\usepackage{algorithm}
\usepackage{algpseudocode}
\usepackage{caption}
\usepackage{subcaption}
\usepackage{multirow}
\usepackage{amsmath}
\usepackage{hyperref}
\usepackage{comment}
\usepackage{xcolor}

\usepackage{lineno}

\begin{document}




\title{A Reinforcement Learning Approach to Dairy Farm Battery Management using Q Learning}


\author[inst1]{Nawazish Ali*}
\author[inst1]{Abdul Wahid}
\author[inst2]{Rachael Shaw}
\author[inst1]{Karl Mason}

\affiliation[inst1]{organization={School of Computer Science},
            addressline={University of Galway}, 
            city={Galway},
            postcode={H91 FYH2}, 
            country={Ireland}}


\affiliation[inst2]{organization={Atlantic Technological University},
            addressline={Galway}, 
            postcode={H91 T8NW}, 
            country={Ireland}}

\begin{abstract}
Dairy farming consumes a significant amount of energy, making it an energy-intensive sector within agriculture. Integrating renewable energy generation into dairy farming could help address this challenge. However, fluctuations in renewable generation pose a challenge to this integration. Effective battery management techniques are needed to store and utilize the energy generated from renewable sources. The objective of this research is to leverage Reinforcement Learning to develop an effective approach for battery management systems in dairy farming. Our work contributes by implementing a Q-learning algorithm for dairy farm battery management, incorporating wind and solar energy, exploring the state space of the algorithm, and considering Ireland as a case study as it works towards attaining its 2030 energy strategy centered on the utilization of renewable sources. The findings show that the proposed algorithm reduces the cost of imported electricity from the grid by 13.41\%, 24.49\% when utilizing wind generation, and peak demand by 2\%. These findings highlight the effectiveness of Reinforcement Learning for battery management in the dairy farming sector.
\end{abstract}




\maketitle

\section{Introduction}

The growth in the global population has led to an increased demand for food products. Milk maintains an important place in the dietary patterns of individuals across the globe because of its essential nutritious components. According to the Food and Agriculture Organisation (FAO), global milk production rose from 735 million metric tonnes in 2000 to 855 million metric tonnes in 2019 \cite{b2}. The rising demand for dairy products has led to an increase in the number of dairy farms\cite{bib_oecd2020dairy}. These farms heavily rely on electricity for multiple activities like milk cooling, water heating, pumping, and lighting \cite{b3}. Meeting these energy needs requires substantial imports of electricity from external grids. However, the rising cost of electricity necessitates considering alternative energy sources like solar photovoltaic and wind turbines. Embracing renewable energy sources can help satisfy the energy requirements of farms and decrease dependence on the external grid for importing electricity \cite{b4}. By 2030 Irish government aims to transition to a low-carbon economy within the EU, emphasizing renewable energy, secure electricity supply, and enhanced energy efficiency \cite{energy}. This research supports these goals by efficiently managing energy storage, increasing renewable energy use, and reducing carbon emissions by minimizing grid reliance as the grid generates most of the energy by burning fossil fuels \cite{grid_gen}.  

Power generated from renewable energy sources exhibits temporal variability. Employing a battery management system for storing electrical energy is crucial for future use. Different battery management systems are applied to different applications\cite{hannan2021battery}. The use of batteries has the potential to influence the economic aspects of electricity consumption within dairy farming. However, optimizing battery performance necessitates the implementation of different strategies. Various conventional methods have been employed to improve battery efficiency. These approaches include such as Maximising Self-Consumption (MSC) and Time of Use (TOU) \cite{b5}.

In recent years, there has been remarkable progress in Artificial Intelligence (AI), largely driven by the data revolution. This advancement has shown immense potential across various fields, yielding promising results\cite{ai_application}. One significant area within AI is Reinforcement Learning (RL), where agents can operate in stochastic environments without explicit knowledge of the environment or predefined decision-making processes. Instead, with established objective functions. Two main algorithms that are particularly prominent in the field of RL are actor-critic learning and Q-learning \cite{acl,b12}. RL agents can effectively learn a policy in diverse domains through these algorithms. This research aims to use RL agents to learn battery management policies efficiently. The primary goal is to optimize the charging/discharging of a battery to maximize renewable utilization and reduce the cost of electricity imported from the grid.

Existing research has highlighted the efficacy of RL in battery management across diverse contexts. However, the integration of RL into the agricultural sector, specifically in dairy farming for battery management, remains relatively unexplored in current literature.
The main contributions of this research are specified below:
\begin{enumerate}
  \item Using Q-learning we present an autonomous learning approach for efficient battery management in dairy farms, and we demonstrate the effectiveness of our algorithm in achieving improved energy efficiency over established baseline algorithms.
  \item In contrast to existing approaches, this work also analyses the influence of incorporating wind energy data on the effectiveness of battery management. 
  \item We evaluate the impact of variations in state space information on the performance of our Q-learning approach. We explore the impact of additional parameters including load, PV, and wind to determine the most optimal solution.
  \item We extend our experimental analysis to also evaluate the Q-Learning algorithm’s performance using data based on case studies in Finland and Ireland, focusing on dairy farm battery management.
\end{enumerate}

The remainder of this paper is structured as follows: \textbf{Section 2} examines both conventional and AI methodologies in battery management. \textbf{Section 3} formulates the research problem and the proposed methodology. \textbf{Section 4} Evaluate the performance of our proposed approach. Finally, \textbf{Section 5} concludes the research, emphasizing the primary contributions of this work.

\section{Background and Related Work}
Numerous researchers have worked on improving battery management to reduce reliance on external power grids and lower electricity import expenses. Surprisingly, the use of RL in battery management within the dairy farming sector has been unexplored. This study employs RL techniques to manage batteries in dairy farming,
to reduce dependence on external power grids. Researchers investigated different methods for efficiently handling battery management, including conventional battery control methods such as rule-based and dynamic programming strategies, as well as AI methods, mainly RL. There is a rising interest in utilizing AI, particularly because of the volatile nature of the environment within agent interact. RL agents can adapt to volatile or non-deterministic environments.
\subsection{Reinforcement Learning}
RL is utilized in the research to effectively manage the battery controller to maximize the utilization of renewable generation. RL involves interaction between an agent and its environment to maximize the cumulative reward obtained from the environment through specific actions taken by the agent. The environment can be characterized as a Markov Decision Process (MDP). An MDP comprises a state space denoted by \textbf{\textit{S}}, an action space denoted by \textbf{\textit{A}}, a state change denoted by \(p(s_t+1|s_t; a_t)\) where p represents a probability distribution governing state transitions, and a reward function denoted by \(R: S \times A \rightarrow R\). The agent takes actions at each time interval based on the current state observations. The agent changes behavior by considering the outcomes and feedback from previous actions. A policy determines how the agent acts in the environment denoted as \(\pi\). The function \(\pi\) maps each state in a given environment to a probability distribution of possible actions. The reward from a state is defined as the sum of discounted future rewards, which can be mathematically represented as \(R_t = \sum_{i=t}^{T} \gamma^{(i-t)} r(S_i, A_i)\). In this equation, \(R_t\) defines the reward at time \textbf{\textit{t}}. The symbol \(\gamma\) represents the discount factor, ranging from 0 to 1, and \(\i-t\) illustrates the importance of future rewards as compared to immediate rewards. The rewards are influenced by the actions taken, which are determined by the policy \(\pi\). The goal of RL is to develop a policy that maximizes the expected cumulative reward starting from the initial probability distribution. The aim is to maximize the total reward received from the environment.\\
The expected result of acting in a specific state, while obtaining a certain policy, is calculated using the action-value function. Equation \ref{action_value}, a fundamental component in many RL algorithms, provides a means to evaluate the potential outcome of actions within the framework of the given policy.
\begin{equation}
Q^\pi(s_t, a_t) = E_\pi \left[ \sum_{k=0}^\infty \gamma^k r_{t+k+1} \mid s_t, a_t \right]
\label{action_value}
\end{equation}
Equation \ref{action_value} denotes the action-value function for a given policy \(\pi\) at a specific time t, concerning the state \(s_t\) and action \(a_t\). The returned value represents the expected cumulative discounted reward for taking action at in-state \(s_t\) and then following policy \(\pi\) for all future time steps. The symbol \(E_\pi\) denotes the expected value under policy \(\pi\). Additionally, the summation \(\sum_{k=0}^\infty\) denotes the sum over all potential future time steps that begin from time \(t+1\). The variable \(r_{t+k+1}\) denotes the reward acquired at time t+k+1 subsequent to executing the action \(a_t\) within the state \(s_t\). The \(\gamma\) represents the discount factor which maximizes the future reward.\\
Q-learning is one of the basic RL algorithms that does not require a model of the environment. It is commonly used to determine the best policy for selecting actions in a finite Markov decision process. This approach aims to get information regarding the significance of an action within a specific state, enabling an agent to make decisions that optimize the overall accumulated reward over a given period. The algorithm comprises the process of updating Q-values, which are action-value pairs, that are stored within a table. Each Q-value corresponds to the anticipated utility of executing a specific action within a particular condition, subsequently following the optimal policy. The main formulation to update the Q-value is depicted in the Equation \ref{update_q_value}
\begin{equation}
Q(s_t, a_t) \leftarrow Q(s_t, a_t) + \alpha [r_t + \gamma \max_{a} Q(s_{t+1}, a) - Q(s_t, a_t)]
\label{update_q_value}
\end{equation}
The update rule for the Q-value of the current state-action pair $(s_t, a_t)$ involves a scalar factor $\alpha$, Also known as a learning rate which controls the rate at which the agents will explore the environment ranging from 0 to 1 and scales the difference between the observed reward \textbf{\textit{R$_t$}} plus the discounted estimate of the maximum Q-value for the next state \(s_{t+1}\) (discounted by a factor \(\gamma)\) and the current estimate of the Q-value for the current state-action pair \(Q(s_t, a_t)\).\\

\subsection{Conventional Battery Control Methods}
Many studies have focused on identifying effective operational strategies for PV battery systems, for different objectives \cite{b13}. Specifically, the Maximising Self-Consumption (MSC) and Time of Use (TOU) methods for battery charging \cite{b14}.

The MSC is a method used for managing battery charging and discharging by maximizing the utilization of solar power generation. It charges the maximum amount of solar energy available \cite{b13}. Braun et al. highlighted that optimal battery usage significantly increases the local consumption of solar energy\cite{b15}. Further, they investigated the optimal sizing of photovoltaic systems \cite{b16} and the capacity requirements for energy storage \cite{b17}, aiming to maximize the use of locally generated solar power and reduce reliance on external power grids. In their comprehensive review, Luthander et al. analyzed previous studies on solar power self-consumption in buildings, concluding that proper battery sizing can improve self-consumption rates by 13-24\% \cite{b18}. Sharma et al. conducted a study on the optimization of battery size for zero-net energy homes equipped with rooftop solar panels in South Australia, employing the MSC operational strategy \cite{b19}. Their findings suggest that installing suitable batteries can enhance the self-consumption of solar energy by 20-50\% \cite{b20}.

TOU strategy uses electricity prices for charging and discharging the battery; it charges the battery when the prices are low and discharges at peak times. Feed-in Tariffs (FiT) and TOU pricing strategies, implemented in several countries, aim to enhance the adoption of Photovoltaic Battery (PVB) systems and encourage consumer involvement in energy management, which has been a significant area of research. \cite{b21}. Other studies have focused on the TOU tariff method for efficient battery management. For instance, Christoph et al. utilized optimization techniques to refine the TOU rate structure \cite{b22}, while Li et al. developed TOU tariffs using the Gaussian Mixture Model \cite{b23}. This approach has enabled prosumers to get economic advantages by taking advantage of FiT and adapting to varying electricity prices during peak and off-peak times, which is a main benefit of the TOU strategy \cite{b24}. Research by Gitizadeh et al. and Hassan et al. explores optimizing battery capacity, by utilizing TOU \cite{b25}\cite{b26}. Additionally, Ratnam et al. found that many PVB system users were able to achieve significant annual cost reductions through FiT programs \cite{b27}.

\subsection{Reinforcement Learning for Energy Management}

RL algorithms are widely used in various applications. Wei et al. implemented dual iterative Q-learning for managing batteries in smart residential settings \cite{b28}. This form of Q-learning is designed to enhance energy management in smart homes by optimizing the charging and discharging of batteries. Similarly, Kim et al. developed an RL-based algorithm for energy management in smart buildings \cite{b29}. Their approach uses RL to dynamically identify the most effective energy regulation strategy based on real-time data. Ruelens et al. also applied RL, but to the operation of an electric water heater \cite{b30}, using the algorithm to boost the heater's energy efficiency by learning and adapting to real-time user demand and grid conditions. Research indicates that RL can significantly enhance both efficiency and cost-effectiveness in energy consumption within smart grids. Furthermore, Li et al. introduced a multi-grid RL method to optimize the energy efficiency and comfort of Heating, Ventilation, and Air Conditioning (HVAC) systems \cite{b31}. This method balances HVAC energy use with maintaining optimal room temperature and humidity. Their findings suggest that this approach effectively optimizes energy consumption while ensuring comfortable indoor environments.

\subsection{Reinforcement Learning for Battery Management}

Numerous studies have explored battery management using RL. Foruzan et al. introduced the use of RL for managing energy in microgrids \cite{b32}. They employed an RL system capable of adapting in real-time to changing energy needs and generating renewable energy, enhancing the energy efficiency of microgrids. RL is effective in improving energy consumption in a cost-efficient manner. In a similar application, Guan et al. developed an RL-based solution for controlling domestic energy storage to reduce electricity cost \cite{b33}. This RL method optimizes the charging and discharging of energy storage systems, helping decrease peak power demands and shift energy usage to cheaper, off-peak times, lowering electricity bills. Their simulations demonstrated that this strategy could effectively reduce the electricity cost associated with household energy storage systems. Liu et al. explored the use of Deep Reinforcement Learning (DRL) for optimizing energy management in households \cite{b34}. This study utilized a DRL system designed to enhance energy efficiency in smart homes by constantly learning the most effective energy management strategies. In simulated smart home environments, this DRL-based approach was more efficient and cost-effective than traditional rule-based methods, indicating its potential to significantly improve energy management in intelligent residential settings. Cao et al. proposed the DRL method for battery charging and discharging, handling power price uncertainty, improving the accuracy of the degradation model, and non-linear charging and discharging efficiency \cite{cao2020deep}. They demonstrated the algorithm's efficacy and performance by testing it on historical wholesale electricity data from the United Kingdom. Yu et al. use Deep Deterministic Policy Gradient (DDPG) for the home energy management system to minimize electricity cost by scheduling HVAC systems and Effective Solutions for Storing (ESSs) \cite{yu2019deep}. By leveraging the dynamic prices, the results demonstrated that the proposed algorithm saves energy cost by 8.1\%-15.21\%. \\
Abedi et al. have created a real-time intelligent battery energy control system for residential buildings that incorporates solar panels, battery energy systems, and grid connectivity by using Q-learning. The results of their study demonstrate that the algorithm effectively decreases the monthly electricity cost by 7.99\% to 3.63\% for house 27 and 6.91\% to 3.26\% for house 387. Wei et al. proposed the DDPG a DRL algorithm for the fast charging of lithium-ion batteries (LIB) \cite{wei2021deep}. They compare the proposed algorithm with the rule base by considering different constraints i.e. LIB temperature, charging rapidity, and degradation. Huang et al. introduce Proximal Policy Optimisation (PPO) as a DRL algorithm to optimize the capacity scheduling of solar battery systems \cite{huang2020deep}. To enhance the safety of the battery, a safety control algorithm is implemented by utilizing a serial approach incorporated with a PPO algorithm. Their findings indicate that the proposed algorithm outperforms other DRL algorithms. Cheng et al. propose a periodic deterministic policy gradient (PDPG) to schedule the charging of multi-battery energy storage systems (MBESS) \cite{cheng2023reinforcement}. Their research shows that compared to the DPG algorithm, the PDPG algorithm reduces power cost by 8.79\%. Paudel et al. employ the MDP framework to efficiently manage battery storage systems' charging and discharging operations by considering the electricity price fluctuations and other relevant parameters \cite{paudel2023deep}. The authors substantiate their method's effectiveness by installing 150 fast charging stations and a battery storage system throughout the Pennsylvania-New Jersey-Maryland region. The studies mentioned above show RL's impact on battery management applications. 

The conventional and RL studies underscore the importance of maximizing local energy utilization and optimizing battery usage. However, some limitations have been identified in these works that our research aims to address. Firstly, they did not address performance variations under diverse weather conditions and geographical locations, besides the impact of fluctuating energy prices and renewable generation. Secondly, these studies focus solely on one renewable source and do not consider the effects of integrating other energy sources. Lastly, all the conventional and RL methods have been applied in smart homes and buildings, but their adaptation to dairy farm battery management remains largely unexplored. Dairy farms typically consume more energy than households or offices due to operational needs and reliance on high-energy equipment like milking machines and milk cooling systems, which account for 20-30\% of the farm’s electricity. Furthermore, research has shown that electricity consumption per dairy cow ranges from 4 to 7.3 kWh/week\cite{b3}. In contrast, households and offices use energy mainly for heating, cooling, lighting, and appliances. The unique requirements of dairy farming operations lead to higher load consumption and diverse consumption patterns.  This research addresses these gaps by demonstrating how a Q-learning algorithm optimizes battery management in dairy farming settings. It also mitigates drawbacks by testing the proposed methodology across various locations and weather conditions, integrating multiple renewable sources, and considering electricity price fluctuations.

\section{Methodology}
\subsection{System Design}

\begin{figure*}
\centerline{\includegraphics[width=\textwidth]{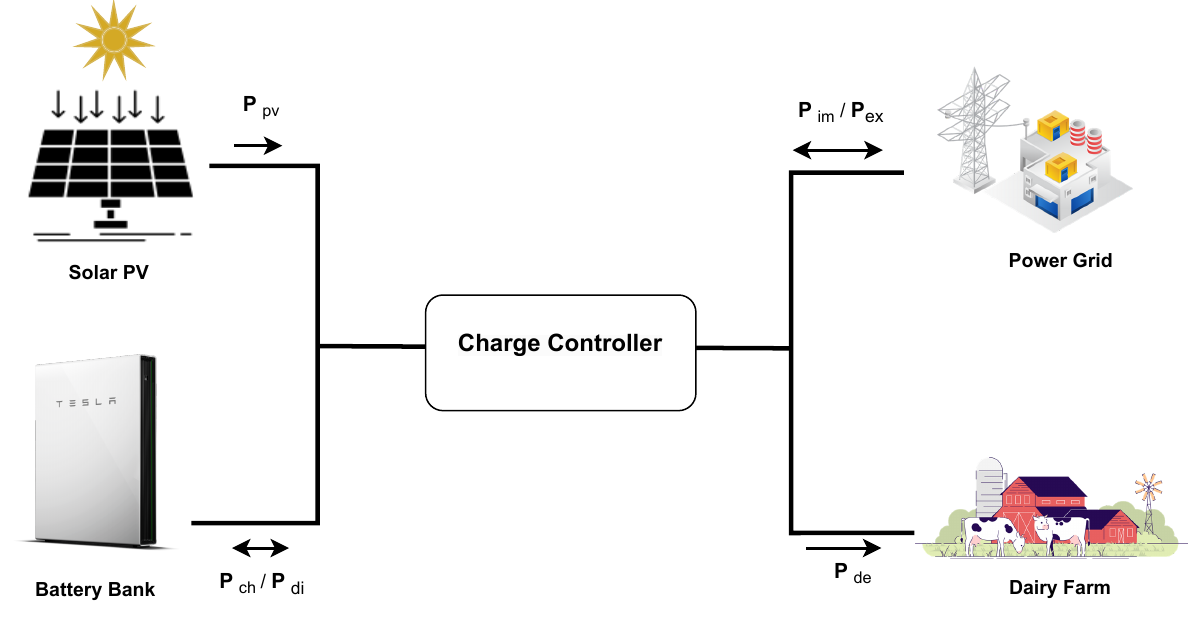}}
\caption{Overview of the system architecture.} \label{design}
\end{figure*}

The PVB system connected to the grid, as depicted in Figure \ref{design}, includes a set of components: solar panels, a battery storage unit, the power grid, and a dairy farm that utilizes electricity from both solar and grid sources.  The energy storage system considered for this research is the Tesla Powerwall 2.0, which offers a substantial capacity of 13.5kWh and supports both charging and discharging 5kW \cite{b36}. The PV-generated electricity is used to meet the farm's load, charge the battery, or sell it back to the grid, according to the operational requirements. The role of the charge/discharge controller is to charge and discharge the battery according to the renewable generation, electricity demand, and price of electricity. Meanwhile, the power grid is connected to the dairy farm and the battery. It supplies electricity when there is high demand and low renewable generation. The battery storage is used to satisfy the farm's extra energy needs, a process commonly referred to as peak shaving. This involves using excess energy demands by utilizing stored power in the battery\cite{hannanpeakshaving}.

\subsection{Data and Price Profile}

\begin{figure}[h]
     \centering
     \begin{subfigure}{0.49\textwidth}
         \centering
         \includegraphics[width=\textwidth]{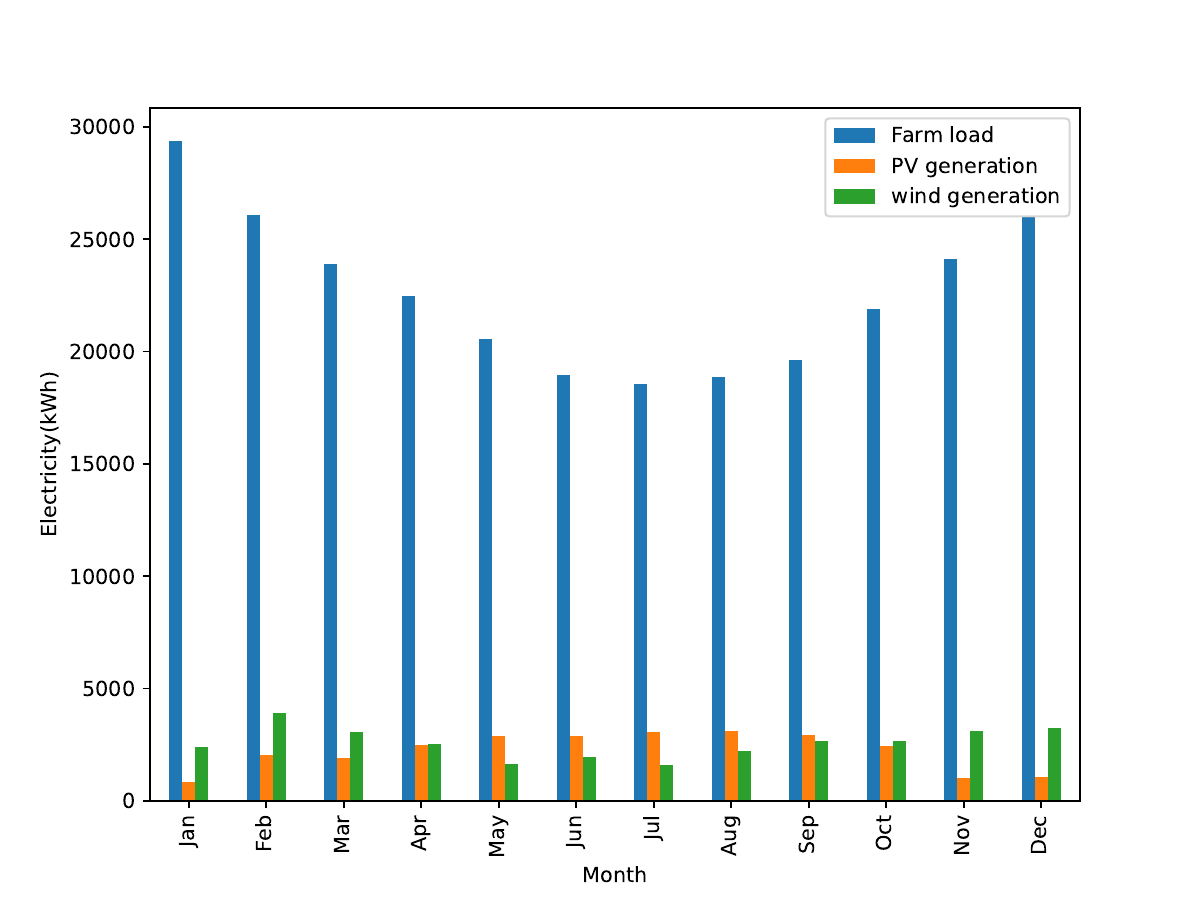}
         \caption{Finland farm load and PV and wind generation for one year.}
         \label{load_data}
     \end{subfigure}
     \hfill
     \begin{subfigure}{0.49\textwidth}
         \centering
         \includegraphics[width=\textwidth]{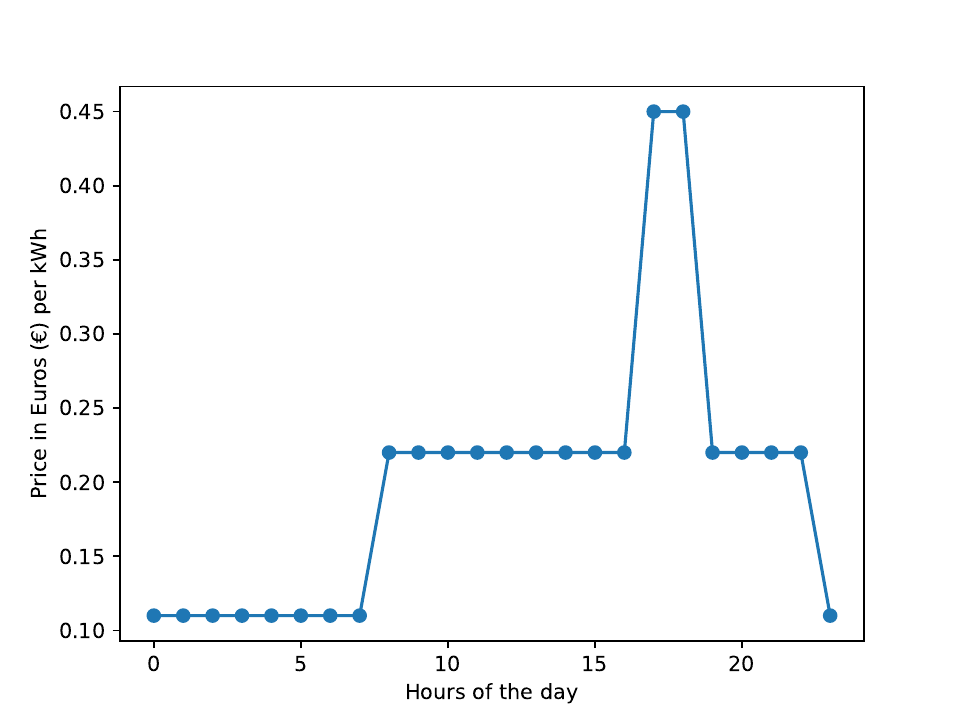}
         \caption{Finland electricity price profile}
         \label{price_data}
     \end{subfigure}
        \caption{Dairy farm electricity, solar photovoltaic generation, and price data from Finland.}
         \label{overall_fin_data}
\end{figure}

For this study, two datasets were used: One dataset from Finland to train the algorithm and a second dataset collected from Ireland to evaluate the performance of the algorithm.  The Finland dataset has information about the load demand from dairy farms, PV generation, wind generation, and electricity prices. The load data is collected from \cite{b10} and provides hourly electricity consumption over a year. Figure \ref{load_data} demonstrates the monthly distribution of electricity demands for a dairy farm and PV generation and wind energy generated by the dairy farm throughout the year. The dataset consists of a dairy farm that has approximately 180 cows and has an estimated annual electricity usage of around 261 megawatt-hours (MWh). The PV and wind data was collected from the System Advisor Model (SAM) having a capacity of 20kW \cite{b35}. The Finland electricity price data was collected from a Helsinki electricity supply company \cite{b_fin_prc}. This price data is dynamic and includes three different price levels \cite{b_tou}. The lowest rate is during off-peak hours, the standard rate applies for most of the day, and a higher peak rate is charged during the busiest hours. Specifically, the pricing is segmented into three time periods. The off-peak hours, with the lowest rate, are from 11 p.m. to 7 a.m. The standard rate applies during two intervals: from 8 a.m. to 4 p.m. and from 7 p.m. to 10 p.m. The peak rate, which is the highest, is charged between 5 p.m. and 7 p.m. Figure \ref{price_data} shows how these electricity prices fluctuate over the day.

\begin{figure}[h]
     \begin{subfigure}{0.49\textwidth}
         \centering
         \includegraphics[width=\textwidth]{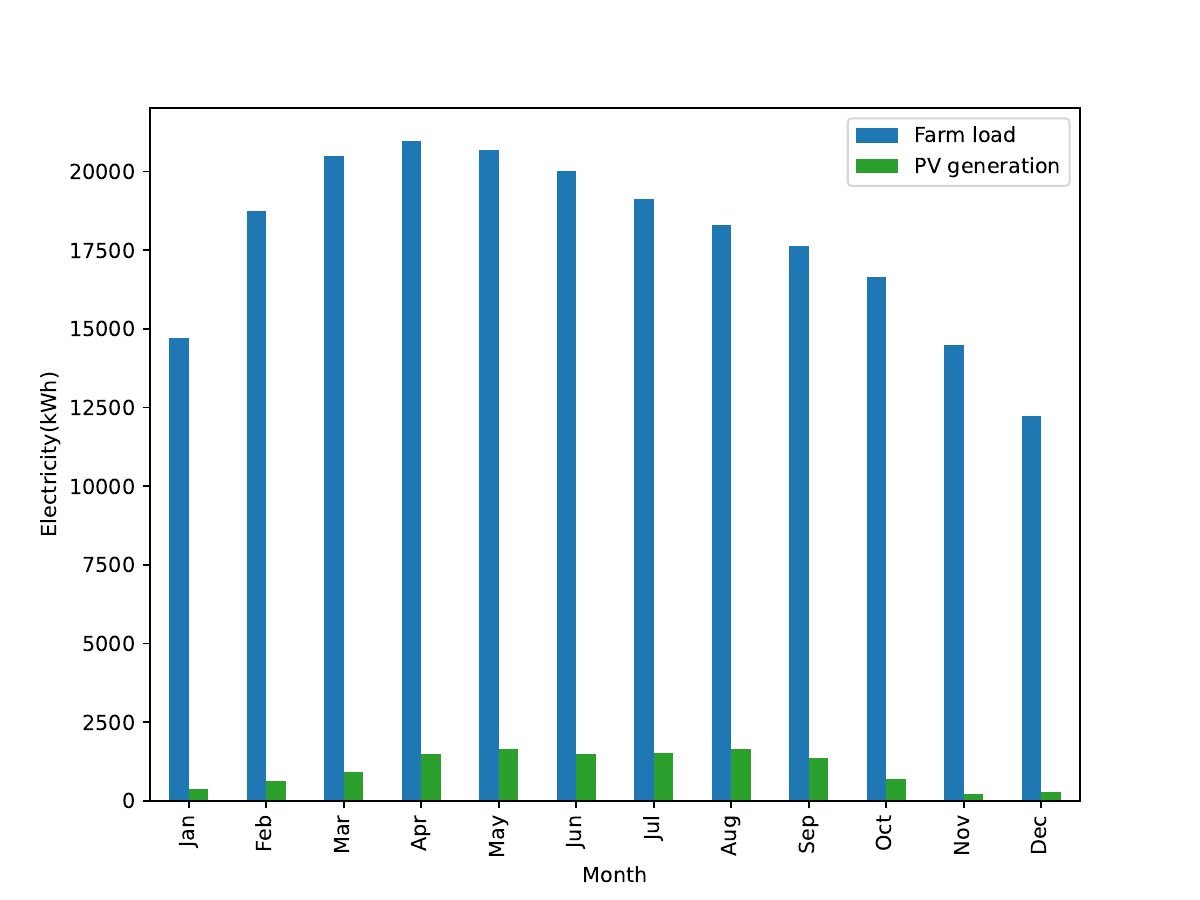}
         \caption{Ireland farm load and PV generation for one year.}
         \label{load_data_ex4}
     \end{subfigure}
     \begin{subfigure}{0.49\textwidth}
         \includegraphics[width=\textwidth]{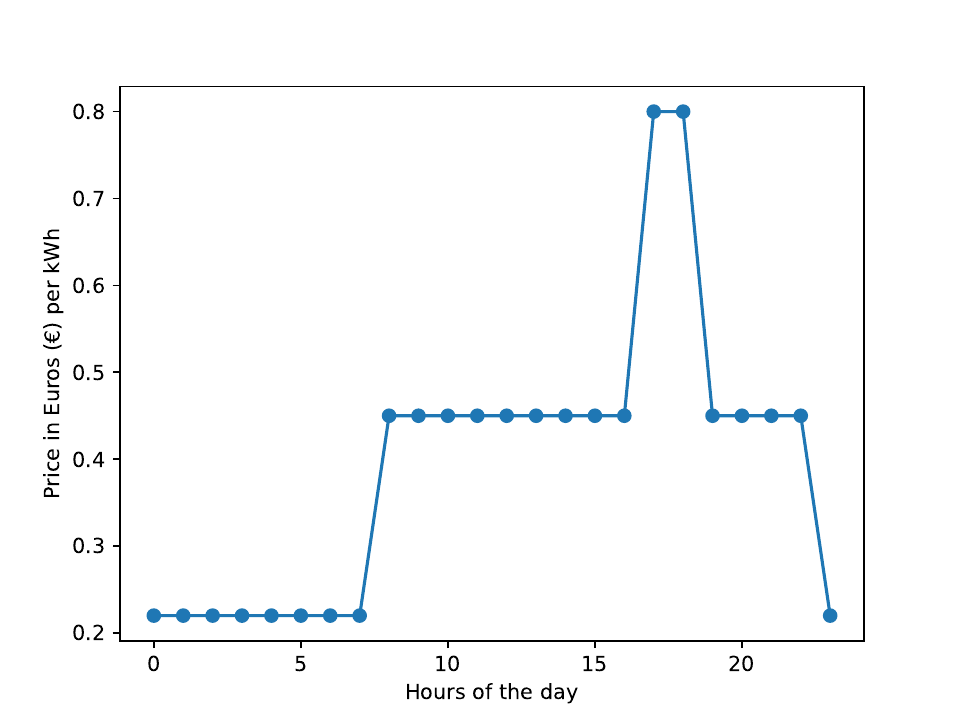}
         \caption{Ireland electricity price profile}
         \label{price_data_ex4}
     \end{subfigure}
        \caption{Dairy farm electricity, solar photovoltaic generation, and price data from Ireland.}
         \label{overall_ir_data}
\end{figure}

The Ireland dataset includes data on farm load, PV generation, and electricity pricing. However, it lacks wind generation data as we were unable to find wind generation data for Ireland. The load consumption data, detailing electricity from the dairy farm over a year, was collected from a study on Irish dairy farms \cite{b_hu}. The PV generation data was collected from SAM\cite{b35} having a capacity of 20kW. The price data is collected from the Ireland electricity supply company Electric Ireland \cite{b_ir_prc}. Figure \ref{overall_ir_data} shows the Irish dairy farm energy consumption and photovoltaic (PV) energy generation and electricity price. This figure illustrates the variations in PV generation and electricity price, it also demonstrates the farm electricity demand patterns. The aim is to explore the relationship between energy consumption and PV generation, particularly in the Irish dairy farm context. Figure \ref{load_data_ex4} specifically illustrates the monthly load demand and PV generation of the dairy farm over one year, while Figure \ref{price_data_ex4} illustrates the price variations over the day.

\subsection{Baseline Battery Controllers}
The battery management system was optimized through the implementation of two rule-based strategies which are MSC and TOU in the baseline algorithm \cite{b5}. The  MSC is a means of optimizing the utilization of surplus energy generated by the PV system through its storage in a battery. The  TOU involved modifying the battery charging process in response to variations in electricity prices. These two strategies were implemented to manage the battery as a baseline comparison method.\\
The MSC strategy is a prevalent energy management approach utilized in PV-integrated energy systems. Its primary objective is to optimize the utilization of PV-generated power for load demand and battery charging. The core principle of this system is that when the energy produced by PV sources surpasses the current energy needs, any excess energy is stored in the battery, and the remaining energy is transmitted to the grid. In cases where the PV generation falls short of the required load, the battery is utilized as the primary source for meeting the load demand. It discharges first to ensure that the load demand is met. If the load demand exceeds the combined capacity of the PV system and battery, external electricity will be purchased from the power grid to compensate for the shortfall. The MSC mostly depends on PV generation and if the PV is not available this strategy doesn't work well. The pseudocode of the MSC strategy is presented in the Algorithm \ref{msc}.

\begin{algorithm}
\caption{MSC Strategy for Battery Management}
\label{msc}
\begin{algorithmic}[1]

\State Initialize $pv\_generation$, $load\_demand$, $battery\_capacity$, $total\_episodes$
\While{$episode = 1$ \textbf{to} $total\_episodes$}
    \State Update $pv\_generation$ and $load\_demand$
    \If{$pv\_generation > load\_demand$}
        \State $excess\_energy \gets pv\_generation - load\_demand$
        \If{$battery\_capacity$ can store $excess\_energy$}
            \State Store $excess\_energy$ in battery
        \Else
            \State Store in battery up to $battery\_capacity$
            \State Transmit remaining $excess\_energy$ to grid
        \EndIf
    \ElsIf{$pv\_generation < load\_demand$}
        \State Use battery to meet $load\_demand$
    \EndIf
\EndWhile

\end{algorithmic}
\end{algorithm}

The adoption of the TOU strategy is aimed at achieving economic gains through the utilization of the price variation between peak and off-peak electricity rates. The primary objective of the TOU strategy charge the battery during the valley price period and subsequently discharge the stored electricity to meet load demand during high/peak periods. In addition, the TOU strategy charges the battery at the highest possible rate from the grid during the off-peak period (23:00-7:00 the following day). In instances of peak prices, the battery is discharged to fulfill the energy demand of the farm when load demand exceeds the capacity of the photovoltaic generation. The pseudocode of the TOU strategy is presented in the Algorithm \ref{tou}.

\begin{algorithm}
\caption{TOU Strategy for Battery Management}
\label{tou}
\begin{algorithmic}[1]

\State Define $peak\_hours, off\_peak\_hours, total\_episodes$
\State Initialize $pv\_generation$, $load\_demand$, $battery\_capacity$, $total\_episodes$, $electricity\_prices$

\While{$episode = 1$ \textbf{to} $total\_episodes$}
    \State Update $pv\_generation, load\_demand, current\_time$
    \If{In off\_peak\_hours and battery not full}
        \State Charge battery from grid at max rate
    \EndIf
    \If{$pv\_generation > load\_demand$}
        \State Store excess PV in battery
    \EndIf
    \If{In $peak\_hours$ and $load\_demand > pv\_generation$}
        \State Use battery to meet shortfall
    \EndIf
\EndWhile

\end{algorithmic}
\end{algorithm}

\subsection{Q Learning}
This paper utilizes the Q-learning approach which is an effective RL technique for efficient battery management used by various researchers as described in the literature. The Q-learning algorithm operates by choosing the action that corresponds to the maximum Q-value in each state. Equation \ref{max_q_value} illustrates the maximum Q-value selection strategy. 

\begin{equation}
Q^*(s_t, a_t) = {argmax}_{a \in A} Q^\pi(s_t, a_t)
\label{max_q_value}
\end{equation}

The symbol \(Q^*(s_t, a_t)\) denotes the optimal action that maximizes the action-value function \(Q^\pi(s_t, a_t)\) at time \(t\), with respect to the state \(s_t\). The mathematical symbol \({argmax}_{a \in A}\) denotes the maximum value of the action-value function across the set of all possible actions belonging to the action space \(A\), given the state \(s_t\). The aforementioned statement implies that the optimal value of the action, denoted by \(Q^*(s_t, a_t)\), results in the maximum reward for the agent in the state \(s_t\).\\
Q-learning algorithms employ the Bellman equation \cite{b_bellman} to choose maximum Q-values and the generalized Bellman equation is expressed in Equation \ref{bellman}.

\begin{equation}
Q_\pi(s, a) = \sum_{s', r} p(s', r | s, a) [r + \gamma \sum_{a'} \pi(a' | s') Q_\pi(s', a')]
\label{bellman}
\end{equation}

Equation \ref{bellman} presents the correlation between the action-value function \(Q_\pi\), the reward, and transition probabilities of the environment. The statement specifies that the value of \(Q_\pi(s, a)\) is equivalent to the summation of the probability \(p(s', r | s, a)\) of transitioning to state \textbf{\textit{S$'$}} and receiving reward \textbf{\textit{R}}, multiplied by the summation of the immediate reward \textbf{\textit{R}} and the discounted value of the subsequent state \textbf{\textit{S$'$}} under the policy \(\pi\), considering all feasible next states \textbf{\textit{S$'$}} and rewards \textbf{\textit{R}}. The parameter \(\gamma\), commonly referred to as the discount factor, plays a crucial role in determining the relative significance of rewards that are obtained immediately versus those that are obtained in the future. The Bellman equation is a fundamental concept within the field of RL, used used for numerous algorithms that aim to acquire knowledge regarding the value function and policy optimization.\\
Q-learning involves using the current estimate of \(Q^\pi\) to improve its future predictions by including the known reward value \(r(s_t, a_t)\). Q-learning fundamentally relies on the concept of Temporal Difference (TD) learning \cite{b11}. In this method, the Q-value is updated after performing an action in the state \(S_t\) and observing the resulting reward \(r_t\) which leads to a transition to the next state \(s_{t+1}\). The TD is mathematically represented in the Equation \ref{update_q_value}.

Empirical evidence supports the notion that as the frequency of visits to each state-action pair's Q-value approaches infinity, the learning rate \(\alpha\) exhibits a decreasing trend concerning the time step \textbf{\textit{t}}. As the value of \textbf{\textit{t}} approaches infinity, the function \(Q(s; a)\) approaches the optimal \(Q*(s; a)\) for all possible state-action pairs \cite{b12}. 
In this study, the Q-learning algorithm was utilized to optimize the management of battery charging and discharging operations to reduce the cost of imported electricity from the power grid. The Q-learning algorithm comprises different components, namely the state space denoted as \(S\), the action space represented by \(A\), and the reward function, which is the aggregate cost of electricity denoted as \(R\).

\subsection{Application of Q-Learning to Battery Management}

\begin{figure*}    
\centerline{\includegraphics[width=3in]{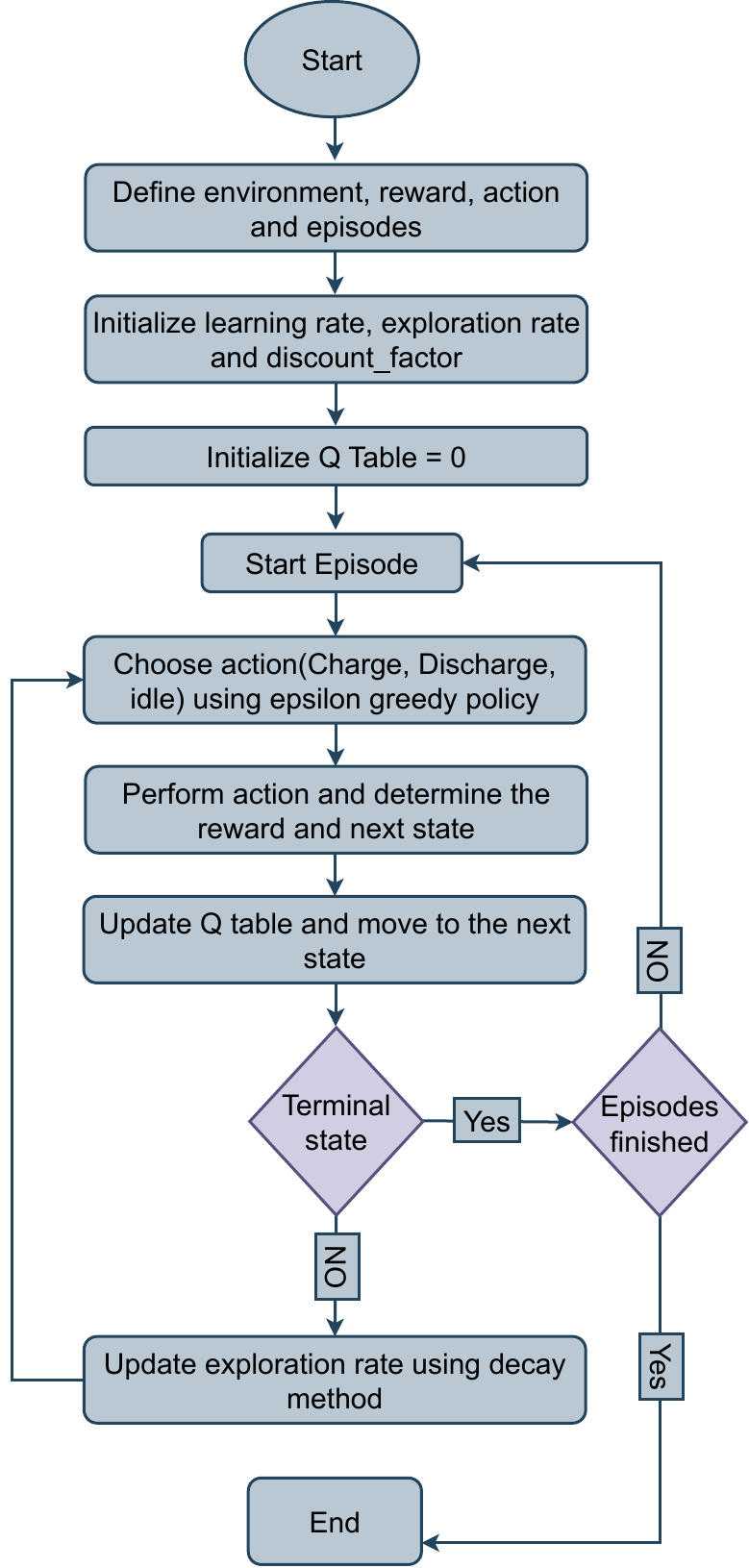}}
\caption{Flow chart of the proposed algorithm.} \label{flow_chart}
\end{figure*}

In this study, Q-learning is employed as a means of effectively managing the process of battery charging and discharging. This is achieved through the exploration of the state space and action space, which are integral components of the environment. The reward is calculated by considering various actions, such as charging, discharging, or remaining idle, in response to factors such as renewable generation and electricity prices. The state space, action space, and reward are explained below. The proposed algorithm is illustrated in the flow chart shown in Figure \ref{flow_chart}. The algorithm begins by initializing the environment, specifying the available actions, defining a strategy for computing rewards, and determining the number of episodes. Subsequently, it initializes the learning rate and exploration rate to 0.8, the discount factor to 0.9, and initializes the Q Table to 0. The algorithm employs the weight decay with the decay of 0.0001 to gradually decrease the learning rate and exploration rate concerning the episodes. To determine the appropriate action, the algorithm uses the epsilon-greedy policy.

\subsubsection{State Space (S)}
This study incorporates two state variables, namely the time component denoted as \(hour\) and the battery charge component denoted as \(SOC\). Equation \ref{state} illustrates the state space for the battery management environment.

\begin{equation}
S = \{hour, SOC\}
\label{state}
\end{equation}

The temporal component \(hour\) represents the hour of day which allows the learning agent to learn about dairy farm load consumption and PV energy generation. \(SOC\) represents battery State of Charge (SOC) controllability. In this study, SOC was divided into ten bins, ranging from 0 to 9. Each bin corresponds to a 10\% increment of the battery charge, effectively discretizing the state space of the battery management environment. This approach is taken to ensure that the distribution of SOC is evenly distributed and simplifies the complexity of the environment, making it more effective for analysis. SOC is represented as \(SOC = SOC_c/SOC_{max}\). The \(SOC_c\) represents the battery charge at the current timestamp and \(SOC_{max}\) represents the battery maximum capacity.

\subsubsection{Action Space (A)}
This study examines a set of three actions, namely charging, discharging, or remaining idle, represented as \(A = \{ {charge, discharge, idle} \}\), where an action \(A = charge\), representing the charging of the battery using PV, and from the local utility grid. If \(A = discharge\) discharge the battery when necessary to meet some or all of the energy requirements. In cases where the energy provided by the PV system and the battery is insufficient, it may be necessary to purchase additional power from the grid. If \(A = idle\), the battery is in an idle state and the dairy farm is powered via solar PV and the grid. For selecting the action the Epsilon greedy policy is used.

In reinforcement learning, the epsilon-greedy policy is an approach that is often used with the Q-learning algorithm. It is the policy that helps the agent select an action in a specific state by using exploration and exploitation methods. In exploration, the agent chooses an action randomly without using previous knowledge, but in exploitation, the agent chooses the action using previous knowledge. The agent decides on the exploration based on the value of $\epsilon$, which ranges from 0 to 1. If $\epsilon$ = 0.1, then there is a 10\% chance that the agent will explore the state and take random action on that state.

\subsubsection{Reward (\(R\))}

The reward function, denoted as \(R\), is computed as the cost of electricity imported from the grid and the electricity price at that hour. Equation \ref{reward} represents the detailed mathematical formulation to calculate the reward for the battery management environment

\begin{equation} \label{reward} 
R = \begin{cases}

    -((P_{dem} + (\beta - P_{pv})) \times P_{e}) - Penalty_{charge}   \text{ if } A = charge \\
    -((P_{dem} - P_{pv}) - \gamma)) \times P_{e}) - Penalty_{discharge}   \text{ if } A = discharge \\
    -((P_{dem} - P_{pv}) \times P_{e}) - Penalty_{idle}   \text{ if } A = idle \\

\end{cases}
\end{equation}

In Equation \ref{reward} $R$ represents the reward obtained at time $t$, $P_{t}$ denotes the total power generated by the solar panels at time $t$, $P_{dem}$ represents the demand of the electricity by the dairy farm,  $\beta$  represent the charge rate at which battery is charged. $\gamma$ represents the discharge rate at which the battery is discharged in kW, and $A$ represents the action taken at time $t$, which can be either charge, discharge, or idle. $P_{e}$ represents the price of electricity at the current time. The $Penalty_{charge}$ represents the penalty amount by which the agent is penalized if it charges the battery under certain rules,  $Penalty_{discharge}$ is the penalized amount when the agent selects an action to discharge the battery, and $Penalty_{idle}$ is the amount of penalty when agent selects an action idle. The formulations of the penalized terms, which are applied based on the actions taken by the agent, are outlined in Equations \ref{charge}, \ref{discharge}, and \ref{wait}.

\begin{equation} \label{charge} 
Penalty_{charge} = \begin{cases}
    -15 \text{ if } SOC_c \geq SOC_{\text{max}} \text{ and } \text{hour} == \text{peak hours} \\
    -10 \text{ if } SOC_c \geq SOC_{\text{max}} \\
    -10 \text{ if } \text{hour} == \text{peak hours} \\
    +5 \text{ if } \text{hour} == \text{off-peak hours} \\
    
\end{cases}
\end{equation}

Equation \ref{charge} explains how penalties are calculated when an agent chooses an action charge. This penalty depends on the battery's current state of charge $(SOC_c)$ and the time of day. If the agent charges an already full battery $(SOC_{max})$, it gets penalized. A penalty of -15 is applied if the agent charges during peak electricity price hours and the battery is fully charged. The agent is penalized a penalty of -10 in two scenarios: first, if it charges the battery during off-peak hours when the battery is already fully charged, and second if it charges the battery during peak electricity hours.  Contrarily, the agent gets a penalty of +5 for favorable actions like charging the battery at night when electricity prices are lower.

\begin{equation} \label{discharge} 
Penalty_{discharge} = \begin{cases}
   -10 \text{ if } SOC_c \leq SOC_{\text{min}} \\
    -5 \text{ if } \text{hour} == \text{off-peak hours} \\
    +5 \text{ if } \text{hour} == \text{peak hours} \text{ and } \text{ if } SOC_c > SOC_{\text{min}}\\

\end{cases}
\end{equation}

Equation \ref{discharge} explains how penalties are calculated for discharging the battery. This penalty varies based on the battery's current state of charge $(SOC_c)$ and the time of day. The agent faces a penalty of -10 if it discharges the battery below its minimum charge level $(SOC_{min})$. A penalty of -5 is applied if the battery is discharged during off-peak times. However, discharging during peak hours periods results in a reward of +5 if the battery charge is more than the battery's minimum level.

\begin{equation} \label{wait} 
Penalty_{idle} = \begin{cases}
    -10 \text{ if } SOC_c \geq SOC_{\text{min}} \text{ and } \text{hour} == \text{peak hours} \\
\end{cases}
\end{equation}

Equation \ref{wait} outlines the penalty for the agent when it selects the "Idle" action. This penalty depends on the battery's current state of charge $(SOC_c)$ and the time of day. To encourage more usage of battery power, a penalty of -10 is imposed during peak hours if the battery's charge level is above the minimum level.

The proposed Q-learning algorithm for battery management in dairy farming is presented in Algorithm \ref{algo:qlearning}. 

\begin{algorithm}
\caption{Battery Management using Q-learning}
\label{algo:qlearning}
\begin{algorithmic}[1]

\State Initialize $days$, $hours$, $maxSOC$, $learning\_rate$, $discount\_factor$, $epsilon$, $decay$, $steps\_per\_episode$, $total\_episodes$ 
\State Initialize $actions \gets \{\text{'charge'}, \text{'discharge'}, \text{'idle'}\}$
\State Initialize $Q\_table[hours][maxSOC + 1][\text{len}(actions)] \gets 0$

\For{$episode = 1$ \textbf{to} $totalEpisodes$}
    \State $hour \gets \text{1}$
    \State $SOC \gets \text{random between 1 and 10}$
    \While{$steps\_per\_episode$}
        \State Choose $action$ from $actions$ using $\epsilon$-greedy policy
        \State Take $action$, observe $reward$, $new\_hour$, $new\_SOC$
        \State $Q\_value \gets Q\_table[hour][SOC][\text{index of } action]$
        \State $next\_Q\_value \gets \text{max}(Q\_table[new\_hour][new\_SOC])$
        \State $Q\_table[hour][SOC][\text{index of } action] \gets Q\_value + learning\_rate \times (reward + discount\_factor \times (next\_Q\_value - Q\_value))$
        \State $hour, SOC \gets new\_hour, new\_SOC$
    \EndWhile
    \State $learning\_rate \gets \text{max}(learning\_rate - decay, 0.1)$
    \State $epsilon \gets \text{max}(epsilon - decay, 0.1)$
\EndFor

\end{algorithmic}
\end{algorithm}

Algorithm \ref{algo:qlearning} describes a Q-learning method specifically designed for battery management in dairy farming. It initializes Q-values for each state-action pair and then iterates through one million episodes. Within each episode, the algorithm selects an action based on a policy derived from the Q-values, such as the $\epsilon$-greedy strategy. After choosing an action, it observes the reward and the next state that results from that action. The algorithm then updates the Q-value for the current state-action pair. Then algorithm uses the weight decay method to decrease the exploration and learning rate with respect to the number of episodes and set it to a minimum of 0.1. Finally, it repeats the process for all episodes.

\subsection{Experimental Setup}
This research evaluates the proposed Q-learning algorithm for battery management through a series of experiments. 
\begin{enumerate}
\item Experiment 1 involves testing and training the Q-learning algorithm on the Finland dairy farm electricity data.
\item Experiment 2 incorporates Finland wind data for a more detailed evaluation of the algorithm.
\item Experiment 3 tests the performance of the algorithm by exploring the state space.
\item Experiment 4 applies the algorithm to the Irish dairy farm data.
\end{enumerate}

The goal of these experiments is to assess the algorithm's effectiveness in different scenarios, involving parameter adjustments, data analysis, and comparative studies. These experiments aim to demonstrate the algorithm's robustness and potential for optimizing dairy farm energy use.

\section{Results and Discussion}

\subsection{Q-Learning for Battery management}

\begin{figure}
\centerline{\includegraphics[width=\textwidth]{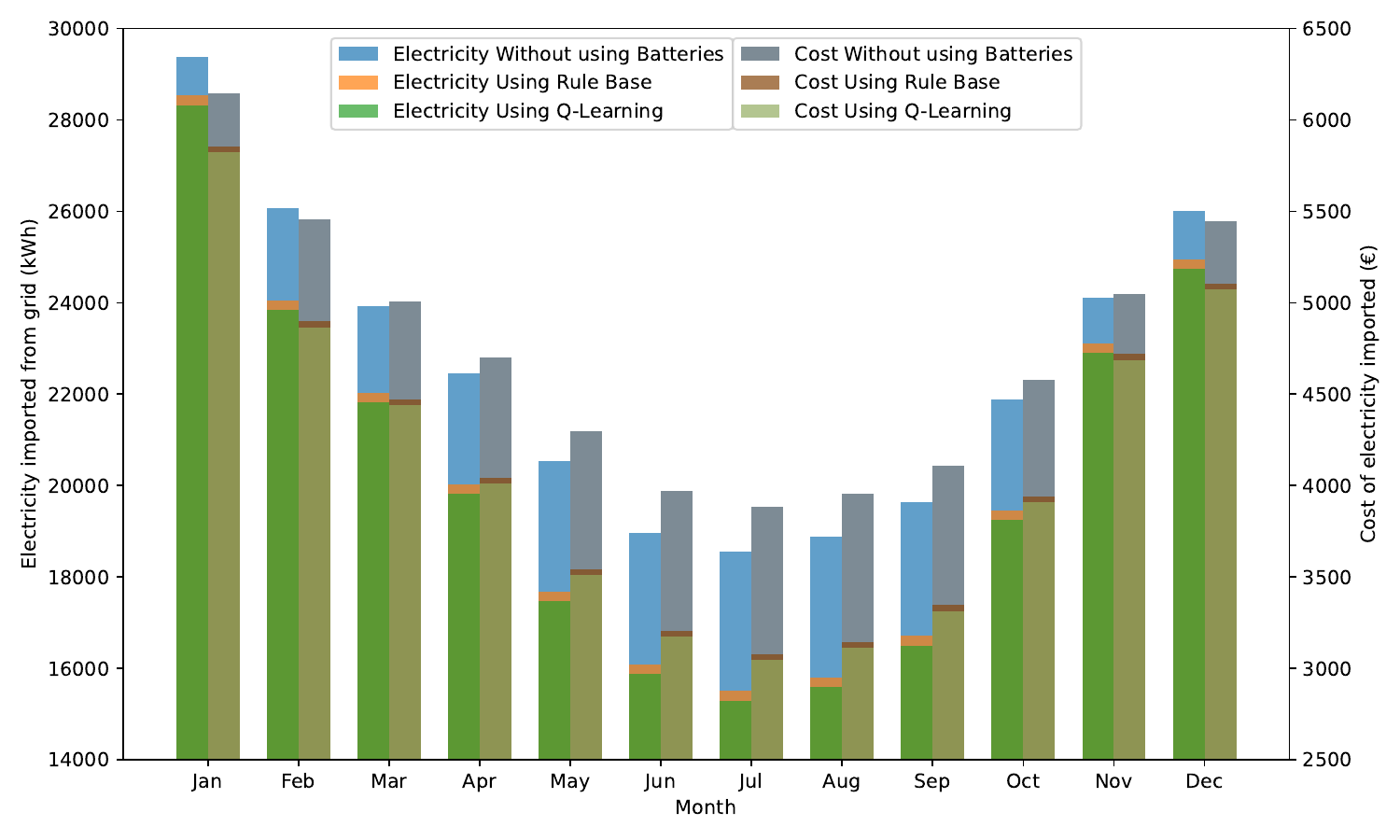}}
\caption{Comparison of the electricity load and cost imported from the power grid by using rule-base and Q-learning on Finland dataset.}
\label{fin_load_price_comparision}
\end{figure}

In this scenario, the Q-learning algorithm was trained to enhance the efficiency of battery management in dairy farming. Its primary goal is to increase the use of PV energy while reducing dependence on the external power grid and to lower energy cost in the dairy farm. This algorithm was trained on one year's data from Finland \cite{b10}. The trained algorithm learned the optimal policy for charging the battery, discharging, and remaining idle, considering state information on battery charge level, time, and energy prices. After training, the algorithm's performance was tested on the same dataset for one year. The findings indicate that the implementation of the Q-learning algorithm decreases the import of electricity by 10.64\%. In comparison, the baseline strategy resulted in a decrease in electricity imports only by 9.72\%. This improvement in the reduction of electricity imports from the grid is presented in Figure \ref{fin_load_price_comparision}, demonstrates the algorithm's effectiveness.

Figure \ref{fin_load_price_comparision} shows a comparison of total electricity imported from the grid and the associated cost of the electricity in each month of the year. The x-axis shows the time in months, while the y-axis on the left side indicates the total electricity imported from the grid while the y-axis on the right side demonstrates the cost of the electricity imported. The graph shows two distinct bars representing: electricity imported from the grid using three methods each marked with a different color; and the cost of the imported electricity by comparing it with three methods, each depicted in a different color. This illustration offers a clear insight into how different energy management strategies affect the overall consumption of electricity and reliance on the grid. The Q-learning effectively reduced electricity imports by 10.64\% and cost by 13.41\% as compared to the baseline algorithm which reduces electricity import by 9.72\% and cost by 12.73\%. These results highlight the effectiveness of Q-learning in optimizing energy usage compared to rule-based battery management and without battery management in reducing grid dependency.

\begin{figure}[h]
\centerline{\includegraphics[width=\textwidth]{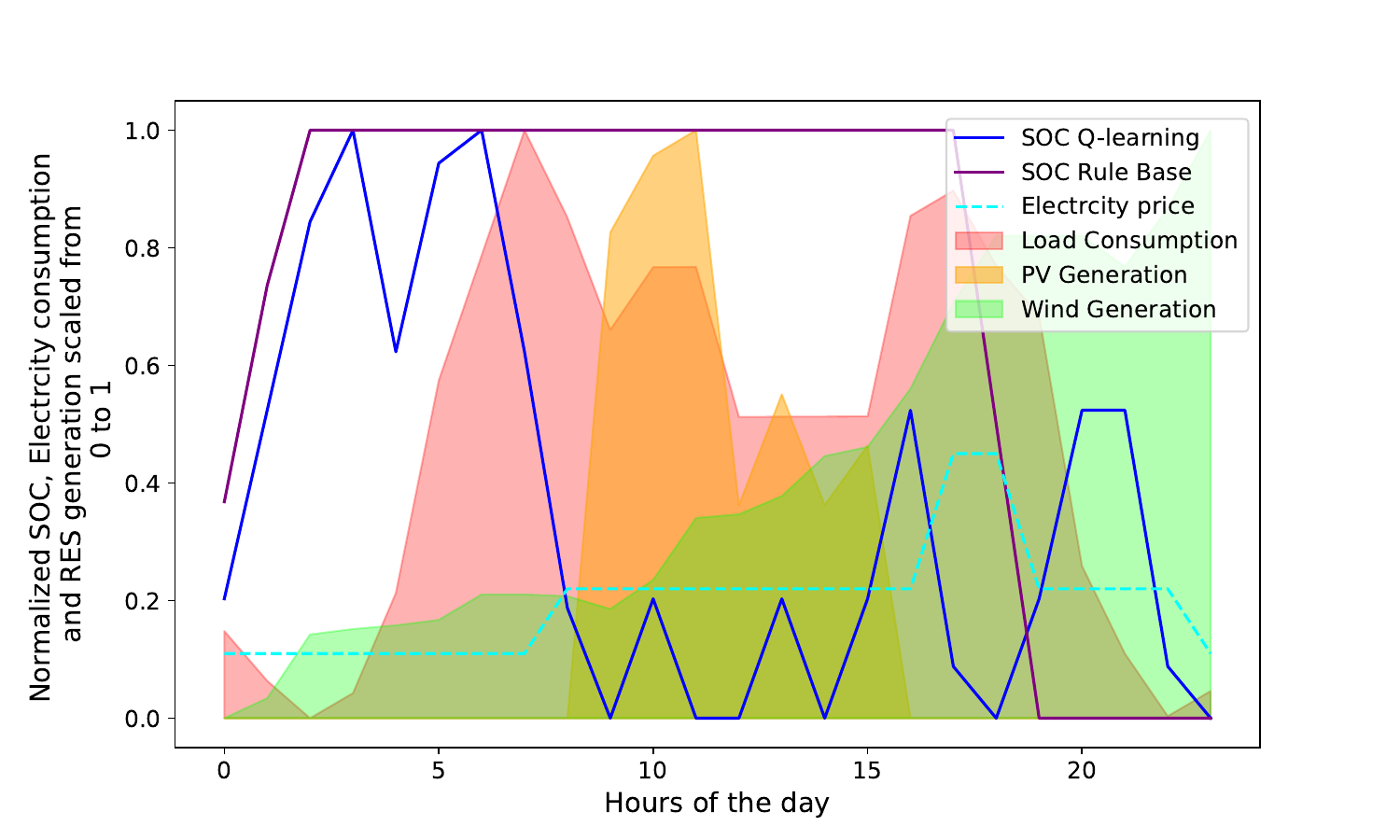}}
\caption{Comparison of the battery charging and discharging by using TOU and Q-learning.}
\label{battery_results}
\end{figure}

Figure \ref{battery_results} illustrates a comparison of battery charging behaviors throughout a day using two methodologies: baseline and Q-learning. Additionally, it displays electricity price, consumption, PV, and wind generation data for the first day of the year. The x-axis represents the hours of the day, while the y-axis indicates the battery and electricity profiles on the farm. This comparison highlights differences in battery charging and discharging behaviors between the two methodologies. The Q-learning method demonstrates enhanced battery management, with results indicating an optimal policy for charging and discharging. Specifically, when it charges the battery during periods of low electricity prices and available PV and wind generation, maximizing the utilization of renewable energy sources. Conversely, during peak hours when electricity prices are high, the Q-learning algorithm discharges the battery.
In contrast, the rule-based method follows a more static approach, based on predetermined rules while Q-learning is adaptive to the current environment. This adaptability allows for more effective optimization of the battery charging and discharging, aligning with fluctuating energy demands and variable PV and wind generation, leading to enhanced efficiency and cost savings for the dairy farm.

\begin{figure}[h]
\centerline{\includegraphics[width=\textwidth]{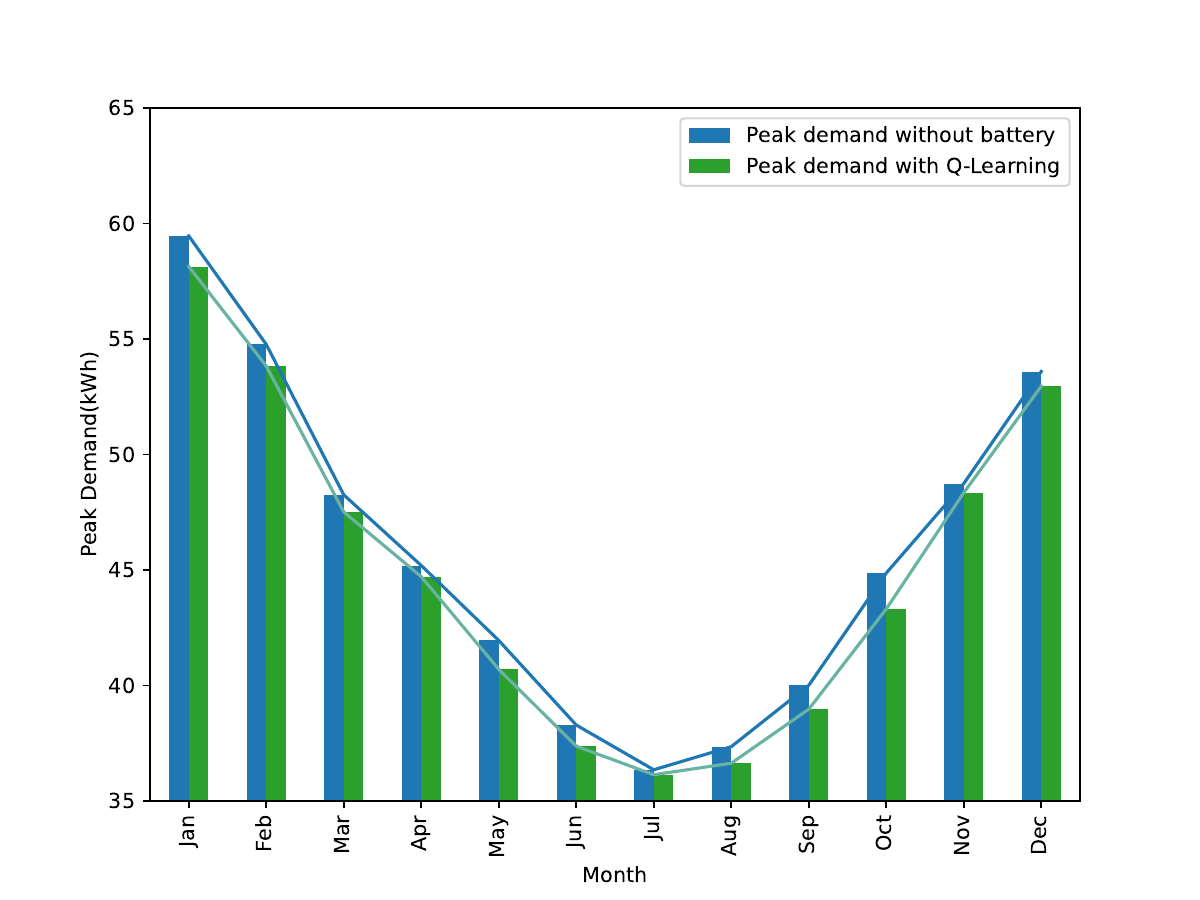}}
\caption{Comparison of the reduction of the peak demand on the grid.}
\label{peak_reduction}
\end{figure}

The peak demand metric is calculated to determine the benefit of Q-learning in terms of its impact on the grid. The algorithm achieved a  2\% reduction in peak demand when using battery management, which is crucial for reducing load from the power grid and reducing the electricity cost in the dairy farm. This reduction, illustrated in Figure \ref{peak_reduction}, compares the peak demand load imported from the grid using Q-learning and without battery management in the month for 1 year, emphasizing the algorithm's effectiveness during periods of peak demand. This significant reduction is particularly important for practical energy management to reduce electricity demand during periods of peak demand.

\subsection{Battery Management with Wind Generation}

\begin{figure}[h]
\centerline{\includegraphics[width=\textwidth]{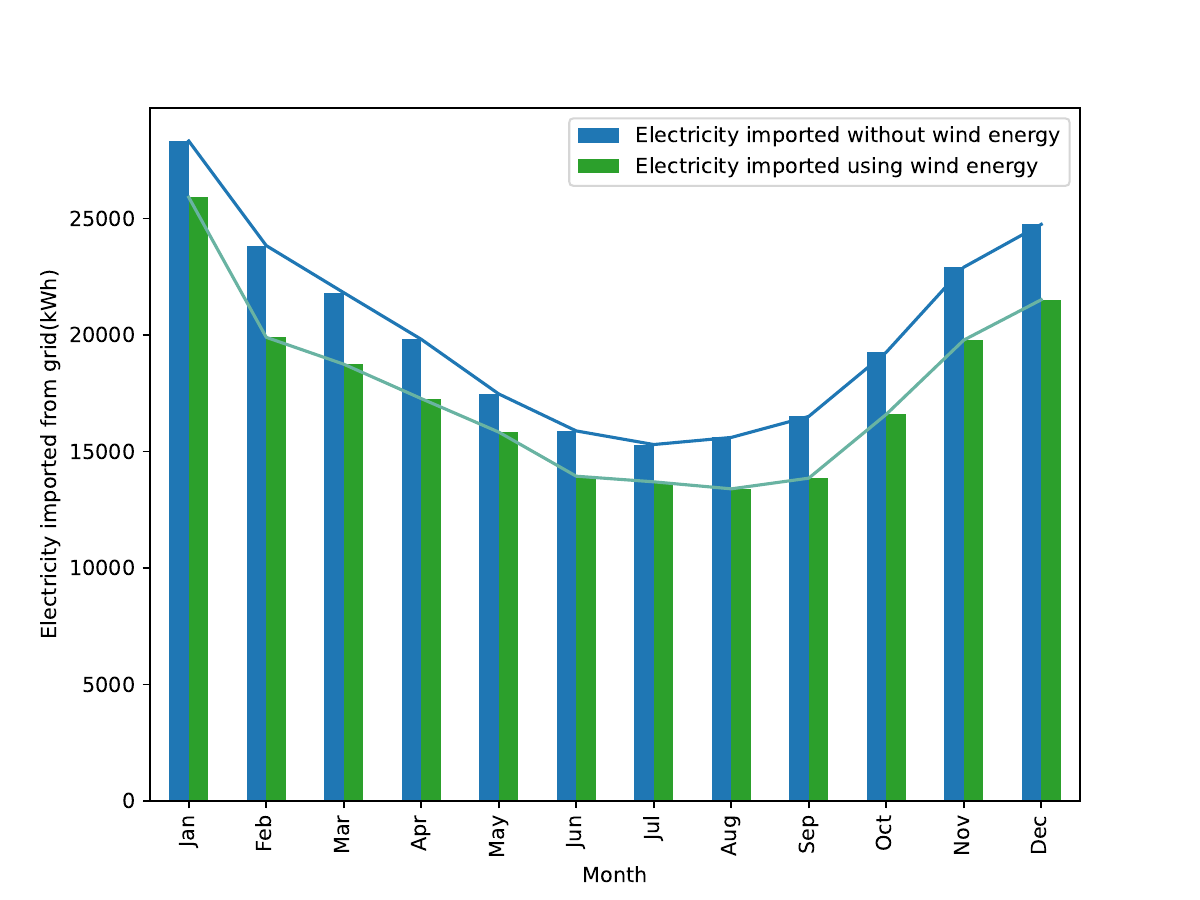}}
\caption{Comparison of the electricity load imported from the power grid by using Q-learning without wind and with wind energy.}
\label{wind_comparision}
\end{figure}

This study investigates the impact of wind energy on the efficacy of the Q-learning algorithm, utilizing the Finland dataset which captures wind generation metrics. The Q-learning algorithm was trained for a total of one million episodes utilizing wind data, in addition to solar data and a load demand from a farm over one year. After training the algorithm performance is evaluated on the data.  The objective of this experiment is to assess the efficacy of Q-learning in energy management by incorporating both wind and solar sources.\\
Figure \ref{wind_comparision}, shows a comparison between the electricity imported from the grid by utilizing wind energy and without wind energy. The x-axis of the figure represents the months of the year, while the vertical axis represents the electricity imported from the grid. The figure shows that the utilization of wind energy resulted in a decrease of 22.14\% in the import of grid electricity, in comparison to 10.64\% generated without wind energy.\\
The findings of the experiment demonstrate that incorporating wind energy through the utilization of the Q-learning algorithm leads to significant reductions in the cost of imported electricity. The reductions using wind energy reduce electricity cost by 24.49\% compared to 13.41\% reduced without wind energy. By integrating wind energy, the algorithm comprehensively reduces electricity import from the grid during the winter period because the wind generation is high in comparison to the PV generation due to wind storms. In summer periods the wind is not too high which affects the performance of the algorithm. The above-mentioned results show the efficiency of the Q-learning algorithm for battery management and the reduction of imported electricity from the grid when wind energy is incorporated.

\subsection{Investigating State Space}

In this experiment, we explore the impact of expanding the state space on the performance of the Q-learning algorithm, initially developed in Experiment 4.1. The state space of the first experiment is depicted in the Equation \ref{state}. The load demand and PV generation were used in the reward function to calculate the reward.

In this investigation, the load and PV generation are incorporated into the state space of the Q-learning algorithm. The purpose of this is to observe how the algorithm's performance is affected when these variables are part of the state space, instead of using them to calculate reward. The formulation of this modification is depicted in Equation \ref{state_modified}.

\begin{equation}
S = \{ {hour, SOC, load, PV} \}
\label{state_modified}
\end{equation}

To further explore the algorithm's adaptability and efficiency, we expanded the state space to include wind data information. Exploration aims to see how dynamic state space affects the adaptability and efficiency of the algorithm. Also, to see how dynamic state space affects the algorithm learning and decision-making capabilities when wind generation data is added to the state space for the battery management system. The state space for this extended approach, incorporating wind data, is presented in Equation \ref{state_wind}.

\begin{equation}
S = \{ {hour, SOC, load, PV, wind} \}
\label{state_wind}
\end{equation}

\begin{figure}[h]
     \centering
     \begin{subfigure}{0.49\textwidth}
         \centering
         \includegraphics[width=\textwidth]{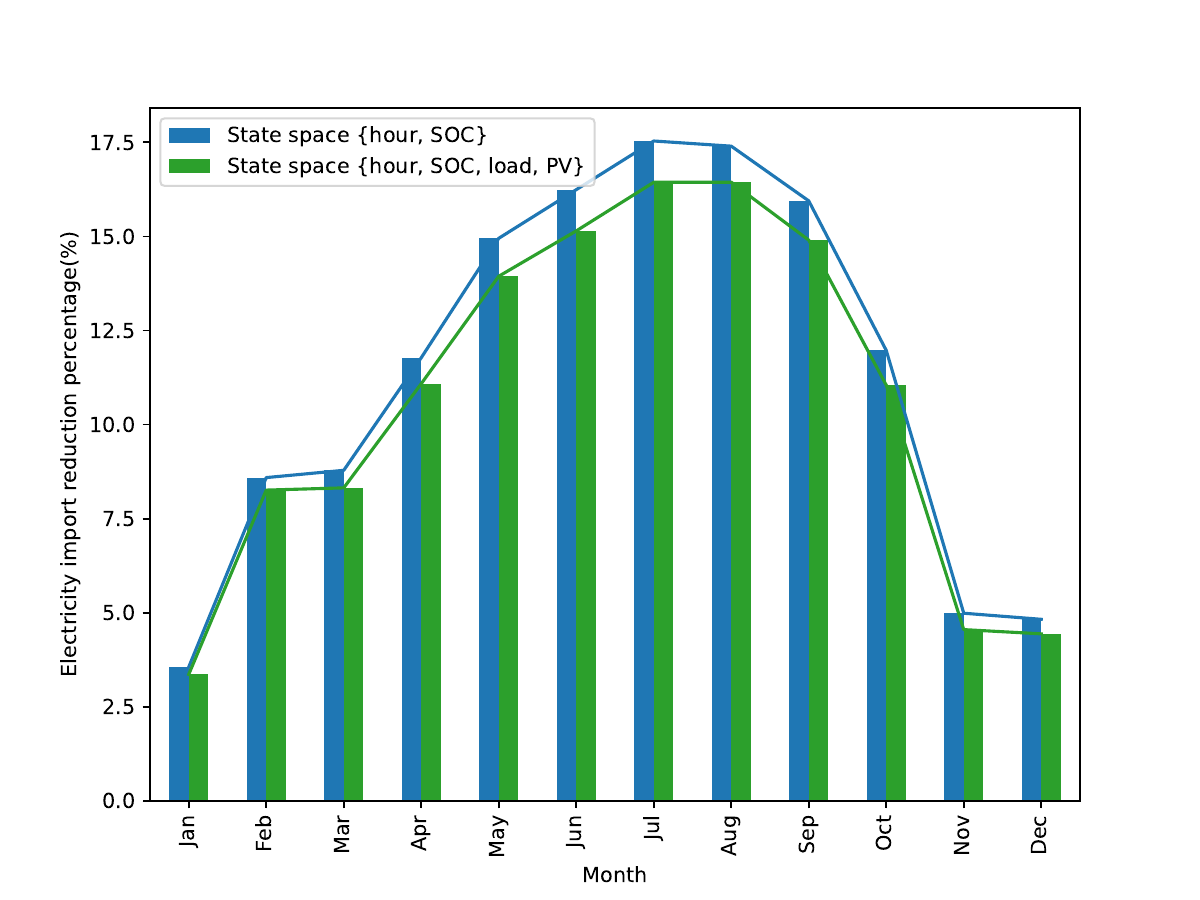}
         \caption{Electricity import reduction without wind}
         \label{state_space_without_wind}
     \end{subfigure}
     \hfill
     \begin{subfigure}{0.49\textwidth}
         \centering
         \includegraphics[width=\textwidth]{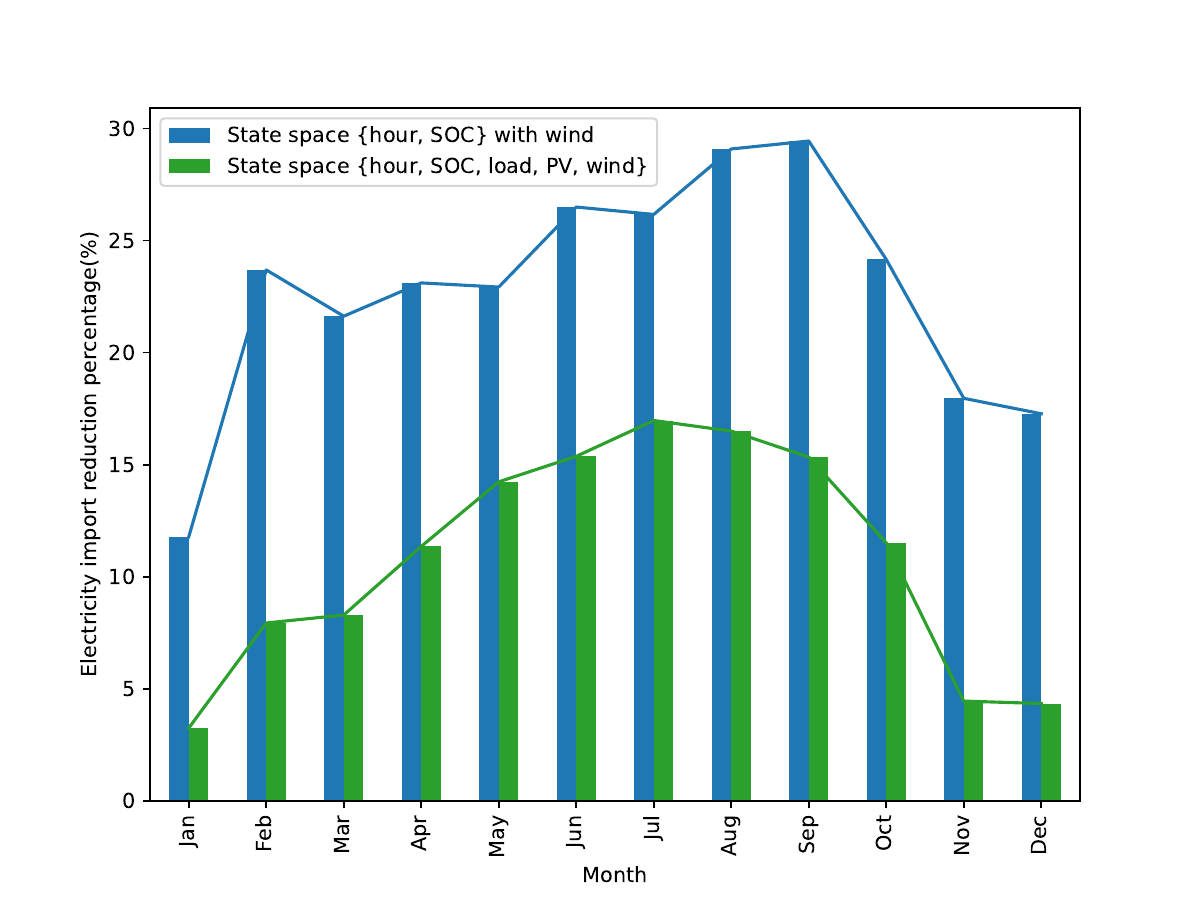}
         \caption{Electricity import reduction with wind}
         \label{state_space_with_wind}
     \end{subfigure}
        \caption{Comparison of electricity import reduction percentage with different state space.}
         \label{state_space_comparsion}
\end{figure}

The Q-learning algorithm is trained and tested with different state spaces including scenarios with and without wind generation data. Figure \ref{state_space_comparsion} compares the algorithm's performance across these different state spaces. Figure \ref{state_space_without_wind} shows how the inclusion of load and PV generation in the state space affects electricity import reduction, compared to the state space from experiment 4.1. We found that the state space from experiment 4.1 is more effective, reducing electricity imports by 10.64\%, compared to the modified state space (with load, and PV) which only achieved a 9.97\% reduction. Figure \ref{state_space_with_wind} illustrates the impact of incorporating wind generation data into the state space. When the state space from experiment 1 is combined with wind data, there is a significant reduction in load import, achieving a 22.14\% decrease. In contrast, a modified state space that includes load, PV, and wind data results in a smaller reduction of only 10.07\%. This shows that expanding the state space adds to the complexity of the environment, which makes it difficult for the Q-learning agent to make optimal decisions. Another reason for this incapability could be due to the curse of dimensionality in Q-learning, where increasing dimensionality leads to sparser data and challenges in achieving expected results\cite{b11}. Additionally, discretizing the state space to manage its dimensionality might have affected performance due to variations in the data.

\subsection{Irish Dairy Farm Case Study}

 In this study, we applied the Q-learning algorithm, originally developed in Experiment 4.1, to the context of Irish dairy farms. The primary objective was to test the algorithm's adaptability using a dataset collected specifically for Ireland. We focused on analyzing electricity consumption and PV energy generation patterns. The main goal of this experiment was to evaluate the efficacy of the Q-learning algorithm in adapting to new data patterns, aiming to optimize battery scheduling and decrease reliance on the electricity grid.

\begin{figure}[h]
\centerline{\includegraphics[width=\textwidth]{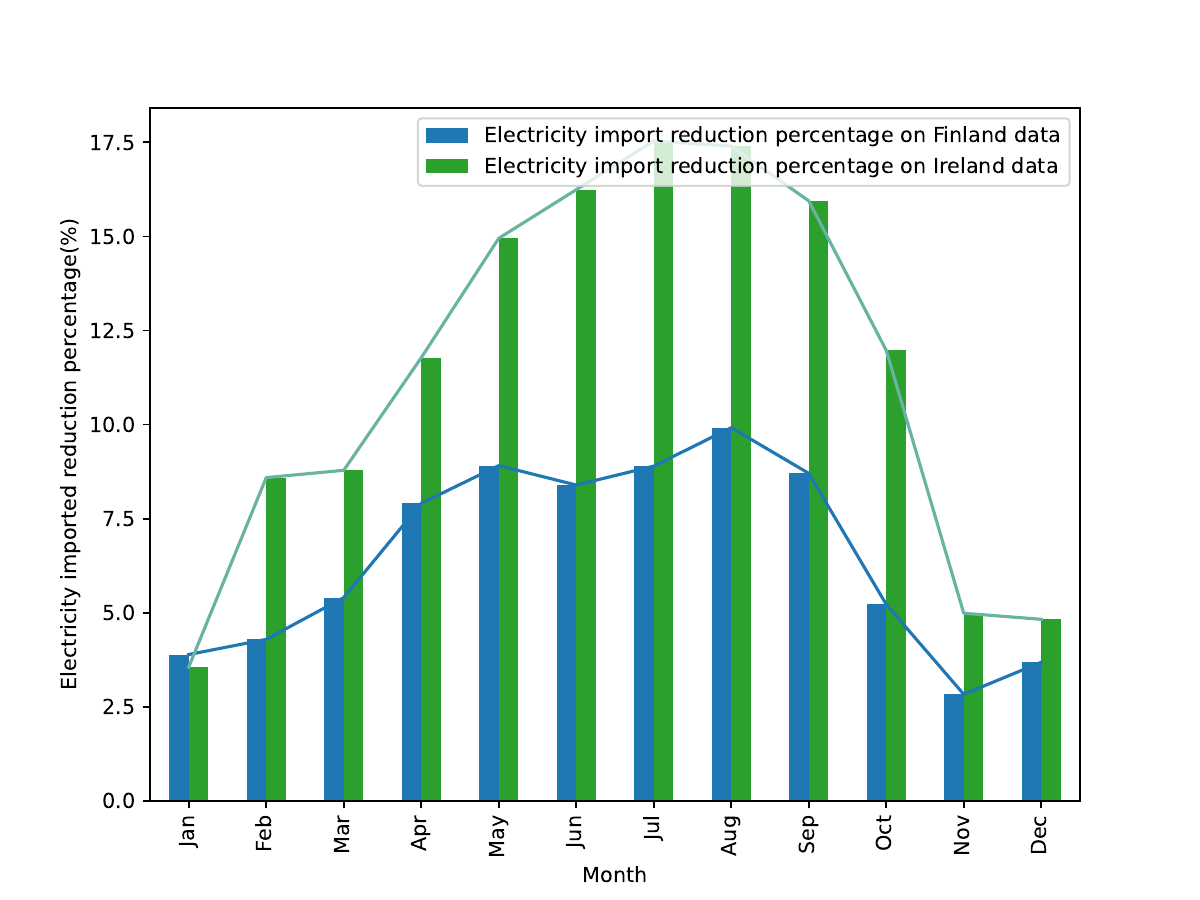}}
\caption{Comparison of the electricity import reduction percentage on Ireland and Finland data.}
\label{ir_fin_comp}
\end{figure}

The comparison of the percentage of electrical load imported from the grid using Q-learning, based on datasets from Finland and Ireland, is illustrated in Figure \ref{ir_fin_comp}. The figure illustrates that the algorithm shows better results in reducing electricity import percentages when applied to the Finland data as compared to the Ireland data. This difference is because the algorithm was trained on the Finland dataset, allowing it to learn and adapt to its specific patterns of electricity consumption and PV generation. In contrast, the Ireland dataset represents a new environment with variations in consumption and generation patterns, which is a new environment for the algorithm in exploring states and deciding on charging and discharging actions. To provide a comprehensive overview of the results of this experiment, we have detailed the results for both Ireland and Finland data in Table \ref{tab_comp}, comparing the proposed algorithm with a baseline algorithm.

Table \ref{tab_comp} shows a comparison of the performance between the baseline method and the Q-learning approach. It highlights the Q-learning algorithm's capability in effectively lowering imported grid load and related cost. Specifically, the Q-learning algorithm reduces electricity import on Ireland data by 6.7\%, an improvement over the baseline's algorithm which reduces by 5.54\%. Additionally, the cost associated with the load was reduced by up to 9.37\% in comparison with the baseline which was reduced by 8.50\%. This comparison showed the adaptability of the Q-learning algorithm in optimizing electricity load and the cost associated with it.

\begin{table}[h]
\caption{Comparison of load and cost reductions for Q-Learning and Rule-Base algorithms on Finland and Ireland datasets.}\label{tab_comp}
\begin{tabular*}{0.9\textwidth}{@{\extracolsep{\fill}}lllll}
\hline
\multirow{2}{*}{Country} & \multicolumn{2}{c}{Rule-based} & \multicolumn{2}{c}{Q-learning} \\ \cline{2-5} 
                         & Load(\%)          & Cost(\%)          & Load(\%)          & Cost(\%)          \\ \hline
Finland                  &  9.72          &  12.73         &  10.64          &  13.41         \\ 
Ireland                  &  5.54          &  8.50         &  6.70          &  9.37         \\ \hline
\end{tabular*}
\end{table}

\section{Conclusion}

In this research, Q-learning is applied to battery management in a dairy farm, using electricity data from Finland. This study involved various experiments to assess the effectiveness of the Q-learning algorithm. This research explored the effect of integrating wind and solar data on battery management and examined how changing the state space of the algorithm impacts its performance. Additional experiments were conducted using data from Ireland to validate the effectiveness of the algorithm. As explained in Section 4, the findings show that the Q-learning algorithm successfully reduced the reliance of the dairy farm on the external grid.

Below are the main findings of this research:
\begin{enumerate}
\item This research utilized Q-learning to manage battery energy in dairy farms, resulting in efficient scheduling of battery loads. The implementation of this strategy resulted in a significant decrease of 13.41\% in the cost of electricity imported from the grid and a reduction in peak demand of  2\%. This shows the proposed strategy's potential to address energy management within the context of dairy farming effectively.

\item The Q-learning algorithm, when applied to wind data integrated with solar data, demonstrated impressive results, achieving a substantial reduction in imported electricity cost by 24.49\%. This emphasizes the algorithm's effectiveness in managing batteries efficiently when wind-generated energy is incorporated with solar energy.

\item Exploring different state spaces in the Q-learning algorithm led to a reduction in electricity import cost. Different experiments were conducted by expanding state space to see the expandability and adaptability of the algorithm.  This improvement highlights the impact of modifying state spaces on battery management in dairy farming when using a Q-learning algorithm.

\item Testing the Q-learning algorithm on the Ireland dataset significantly decreased electricity imports from the grid, with a notable reduction of 6.7\% compared to the 5.54\% achieved with the baseline approach. The outcome shows the Q-learning algorithm's adaptability and effectiveness when applied to data from various regions. 

\end{enumerate}

In the future, we intend to employ DRL algorithms to address the challenge of state space expansion. Deep Learning techniques are well-suited for handling complex problems, and by integrating them, we aim to enhance the model's ability to handle complex state space. This strategy will enhance performance by decreasing dependence on the external grid.

\section*{Acknowledgements} This publication has emanated from research conducted with the financial support of Science Foundation Ireland under Grant number [21/FFP-A/9040].
\bibliographystyle{elsarticle-num}
\bibliography{elsarticle-template-num}

\begin{thebibliography}{10}
\expandafter\ifx\csname url\endcsname\relax
  \def\url#1{\texttt{#1}}\fi
\expandafter\ifx\csname urlprefix\endcsname\relax\def\urlprefix{URL }\fi
\expandafter\ifx\csname href\endcsname\relax
  \def\href#1#2{#2} \def\path#1{#1}\fi

\bibitem{b2}
U.~Nations, \href{https://www.fao.org/home/en/}{Food and agriculture organization of the united nations}, [Online; Accessed 20-April-2023] (2022).
\newline\urlprefix\url{https://www.fao.org/home/en/}

\bibitem{bib_oecd2020dairy}
{OECD}, Dairy and dairy products, \url{https://www.oecd-ilibrary.org/sites/aa3fa6a0-en/index.html?itemId=/content/component/aa3fa6a0-en}, accessed: 2022-11-27 (2020).

\bibitem{b3}
J.~Upton, M.~Murphy, P.~French, P.~Dillon, Dairy farm energy consumption, Dairying: Entering a Decade of Opportunity. Teagasc National Dairy Conference 2010 (2010) 87--97.

\bibitem{b4}
\href{https://ahdb.org.uk/knowledge-library/renewable-energy-opportunities-for-dairy-farmers}{Renewable energy opportunities for dairy farmers}, [Online; Accessed 20-April-2023] (2021).
\newline\urlprefix\url{https://ahdb.org.uk/knowledge-library/renewable-energy-opportunities-for-dairy-farmers}

\bibitem{energy}
Energy, \url{https://www.gov.ie/en/policy/9cd812-energy/}, accessed: 26-Jun-2023 (2020).

\bibitem{grid_gen}
U.~E.~I. Administration, Electricity in the u.s., \url{https://www.eia.gov/energyexplained/electricity/electricity-in-the-us.php}, [Accessed on 26-Jun-2023] (2022).

\bibitem{hannan2021battery}
M.~Hannan, S.~Wali, P.~Ker, M.~Abd~Rahman, M.~Mansor, V.~Ramachandaramurthy, K.~Muttaqi, T.~Mahlia, Z.~Dong, Battery energy-storage system: A review of technologies, optimization objectives, constraints, approaches, and outstanding issues, Journal of Energy Storage 42 (2021) 103023.

\bibitem{b5}
B.~Zou, J.~Peng, S.~Li, Y.~Li, J.~Yan, H.~Yang, Comparative study of the dynamic programming-based and rule-based operation strategies for grid-connected pv-battery systems of office buildings, Applied Energy 305 (2022) 117875.

\bibitem{ai_application}
B.~M. Kumar, V.~Talukdar, H.~Khan, S.~B. Talukdar, A.~Koujalagi, R.~G. Kumar, A.~Gupta, 6 application of al-based, Robotics and Automation in Industry 4.0: Smart Industries and Intelligent Technologies (2024) 110.

\bibitem{acl}
V.~Mnih, A.~P. Badia, M.~Mirza, A.~Graves, T.~Lillicrap, T.~Harley, D.~Silver, K.~Kavukcuoglu, Asynchronous methods for deep reinforcement learning, in: International conference on machine learning, PMLR, 2016, pp. 1928--1937.

\bibitem{b12}
C.~J. Watkins, P.~Dayan, Q-learning, Machine learning 8 (1992) 279--292.

\bibitem{b13}
D.~Azuatalam, K.~Paridari, Y.~Ma, M.~F{\"o}rstl, A.~C. Chapman, G.~Verbi{\v{c}}, Energy management of small-scale pv-battery systems: A systematic review considering practical implementation, computational requirements, quality of input data and battery degradation, Renewable and Sustainable Energy Reviews 112 (2019) 555--570.

\bibitem{b14}
Y.~Zhang, T.~Ma, P.~E. Campana, Y.~Yamaguchi, Y.~Dai, A techno-economic sizing method for grid-connected household photovoltaic battery systems, Applied Energy 269 (2020) 115106.

\bibitem{b15}
M.~Braun, K.~B{\"u}denbender, D.~Magnor, A.~Jossen, Photovoltaic self-consumption in germany: using lithium-ion storage to increase self-consumed photovoltaic energy, in: 24th European photovoltaic solar energy conference (PVSEC), Hamburg, Germany, 2009.

\bibitem{b16}
D.~Talavera, F.~Mu{\~n}oz-Rodriguez, G.~Jimenez-Castillo, C.~Rus-Casas, A new approach to sizing the photovoltaic generator in self-consumption systems based on cost--competitiveness, maximizing direct self-consumption, Renewable energy 130 (2019) 1021--1035.

\bibitem{b17}
N.~J. Vickers, Animal communication: when i’m calling you, will you answer too?, Current biology 27~(14) (2017) R713--R715.

\bibitem{b18}
R.~Luthander, J.~Wid{\'e}n, D.~Nilsson, J.~Palm, Photovoltaic self-consumption in buildings: A review, Applied energy 142 (2015) 80--94.

\bibitem{b19}
V.~Sharma, M.~H. Haque, S.~M. Aziz, Energy cost minimization for net zero energy homes through optimal sizing of battery storage system, Renewable Energy 141 (2019) 278--286.

\bibitem{b20}
E.~Nyholm, J.~Goop, M.~Odenberger, F.~Johnsson, Solar photovoltaic-battery systems in swedish households--self-consumption and self-sufficiency, Applied energy 183 (2016) 148--159.

\bibitem{b21}
L.~Dusonchet, E.~Telaretti, Comparative economic analysis of support policies for solar pv in the most representative eu countries, Renewable and Sustainable Energy Reviews 42 (2015) 986--998.

\bibitem{b22}
C.~M. Flath, An optimization approach for the design of time-of-use rates, in: IECON 2013-39th Annual Conference of the IEEE Industrial Electronics Society, IEEE, 2013, pp. 4727--4732.

\bibitem{b23}
R.~Li, Z.~Wang, C.~Gu, F.~Li, H.~Wu, A novel time-of-use tariff design based on gaussian mixture model, Applied energy 162 (2016) 1530--1536.

\bibitem{b24}
N.~R. Darghouth, R.~H. Wiser, G.~Barbose, Customer economics of residential photovoltaic systems: Sensitivities to changes in wholesale market design and rate structures, Renewable and Sustainable Energy Reviews 54 (2016) 1459--1469.

\bibitem{b25}
M.~Gitizadeh, H.~Fakharzadegan, Battery capacity determination with respect to optimized energy dispatch schedule in grid-connected photovoltaic (pv) systems, Energy 65 (2014) 665--674.

\bibitem{b26}
A.~S. Hassan, L.~Cipcigan, N.~Jenkins, Optimal battery storage operation for pv systems with tariff incentives, Applied Energy 203 (2017) 422--441.

\bibitem{b27}
E.~L. Ratnam, S.~R. Weller, C.~M. Kellett, An optimization-based approach to scheduling residential battery storage with solar pv: Assessing customer benefit, Renewable Energy 75 (2015) 123--134.

\bibitem{b28}
Q.~Wei, D.~Liu, G.~Shi, A novel dual iterative q-learning method for optimal battery management in smart residential environments, IEEE Transactions on Industrial Electronics 62~(4) (2014) 2509--2518.

\bibitem{b29}
S.~Kim, H.~Lim, Reinforcement learning based energy management algorithm for smart energy buildings, Energies 11~(8) (2018) 2010.

\bibitem{b30}
F.~Ruelens, B.~J. Claessens, S.~Quaiyum, B.~De~Schutter, R.~Babu{\v{s}}ka, R.~Belmans, Reinforcement learning applied to an electric water heater: From theory to practice, IEEE Transactions on Smart Grid 9~(4) (2016) 3792--3800.

\bibitem{b31}
B.~Li, L.~Xia, A multi-grid reinforcement learning method for energy conservation and comfort of hvac in buildings, in: 2015 IEEE International Conference on Automation Science and Engineering (CASE), IEEE, 2015, pp. 444--449.

\bibitem{b32}
E.~Foruzan, L.-K. Soh, S.~Asgarpoor, Reinforcement learning approach for optimal distributed energy management in a microgrid, IEEE Transactions on Power Systems 33~(5) (2018) 5749--5758.

\bibitem{b33}
C.~Guan, Y.~Wang, X.~Lin, S.~Nazarian, M.~Pedram, Reinforcement learning-based control of residential energy storage systems for electric bill minimization, in: 2015 12th Annual IEEE Consumer Communications and Networking Conference (CCNC), IEEE, 2015, pp. 637--642.

\bibitem{b34}
Y.~Liu, D.~Zhang, H.~B. Gooi, Optimization strategy based on deep reinforcement learning for home energy management, CSEE Journal of Power and Energy Systems 6~(3) (2020) 572--582.

\bibitem{cao2020deep}
J.~Cao, D.~Harrold, Z.~Fan, T.~Morstyn, D.~Healey, K.~Li, Deep reinforcement learning-based energy storage arbitrage with accurate lithium-ion battery degradation model, IEEE Transactions on Smart Grid 11~(5) (2020) 4513--4521.

\bibitem{yu2019deep}
L.~Yu, W.~Xie, D.~Xie, Y.~Zou, D.~Zhang, Z.~Sun, L.~Zhang, Y.~Zhang, T.~Jiang, Deep reinforcement learning for smart home energy management, IEEE Internet of Things Journal 7~(4) (2019) 2751--2762.

\bibitem{wei2021deep}
Z.~Wei, Z.~Quan, J.~Wu, Y.~Li, J.~Pou, H.~Zhong, Deep deterministic policy gradient-drl enabled multiphysics-constrained fast charging of lithium-ion battery, IEEE Transactions on Industrial Electronics 69~(3) (2021) 2588--2598.

\bibitem{huang2020deep}
B.~Huang, J.~Wang, Deep-reinforcement-learning-based capacity scheduling for pv-battery storage system, IEEE Transactions on Smart Grid 12~(3) (2020) 2272--2283.

\bibitem{cheng2023reinforcement}
G.~Cheng, L.~Dong, X.~Yuan, C.~Sun, Reinforcement learning-based scheduling of multi-battery energy storage system, Journal of Systems Engineering and Electronics 34~(1) (2023) 117--128.

\bibitem{paudel2023deep}
D.~Paudel, T.~K. Das, A deep reinforcement learning approach for power management of battery-assisted fast-charging ev hubs participating in day-ahead and real-time electricity markets, Energy 283 (2023) 129097.

\bibitem{b36}
Tesla.com, How powerwall works, \url{https://www.tesla.com/support/energy/powerwall/learn/how-powerwall-works}, [Online; Accessed 27-March-2023] (2023).

\bibitem{hannanpeakshaving}
M.~Hannan, S.~Wali, P.~Ker, M.~Abd~Rahman, M.~Mansor, V.~Ramachandaramurthy, K.~Muttaqi, T.~Mahlia, Z.~Dong, Battery energy-storage system: A review of technologies, optimization objectives, constraints, approaches, and outstanding issues, Journal of Energy Storage 42 (2021) 103023.

\bibitem{b10}
S.~Uski, E.~Rinne, \href{https://zenodo.org/record/1294967\#.ZFD0Fc7MIQ8}{Data for a dairy farm microgrid solution} (Jun 2018).
\newline\urlprefix\url{https://zenodo.org/record/1294967\#.ZFD0Fc7MIQ8}

\bibitem{b35}
N.~R. E.~L. (NREL), System advisor model (sam), \url{https://sam.nrel.gov}, [Online; Accessed 1-November-2022] (2017).

\bibitem{b_fin_prc}
\href{https://www.helen.fi/en/electricity/electricity-products-and-prices}{Electricity products and prices | helen}, accessed: 2023-11-15.
\newline\urlprefix\url{https://www.helen.fi/en/electricity/electricity-products-and-prices}

\bibitem{b_tou}
{Electric Ireland}, \href{https://www.electricireland.ie/residential/help/smart-electricity-meters/time-of-use-tariffs-for-residential-customers}{Time-of-use tariffs for residential customers}, [Online; accessed 15-November-2022] (2022).
\newline\urlprefix\url{https://www.electricireland.ie/residential/help/smart-electricity-meters/time-of-use-tariffs-for-residential-customers}

\bibitem{b_hu}
K.~H, W.~A, C.~E, M.~K, Modelling electricity consumption in irish dairy farms using agent-based modelling, In Proceedings of the Artificial Intelligence for Sustainability Workshop (AI4S) at ECAI (2023).

\bibitem{b_ir_prc}
{Electric Ireland}, \href{https://www.electricireland.ie/switch/new-customer/price-plans?priceType=P}{New customer price plans}, [Online; accessed 15-November-2022] (2022).
\newline\urlprefix\url{https://www.electricireland.ie/switch/new-customer/price-plans?priceType=P}

\bibitem{b_bellman}
R.~Bellman, Dynamic programming, Science 153~(3731) (1966) 34--37.

\bibitem{b11}
R.~Sutton, Barto:“reinforcement learning: An introduction”, IEEE Trans. Neural Netw 9 (1998) 1054.

\end{thebibliography}

\end{document}